\documentclass[10pt,journal,compsoc]{IEEEtran}
\usepackage{cite}
\usepackage{flushend}
\ifCLASSINFOpdf
  \usepackage[pdftex]{graphicx}
\else
  \usepackage[dvips]{graphicx}
\fi
\usepackage{amsmath}
\usepackage{array}
\usepackage{url}
\usepackage{booktabs}
\usepackage{multirow}
\usepackage[table,xcdraw]{xcolor}

\usepackage{caption}%少标题居中，多则居左

\usepackage{arydshln}%虚线

\usepackage{algorithm} %伪代码
\usepackage{algorithmic}

\begin{document}

\title{Tired of Over-smoothing?\\ Stress Graph Drawing Is All You Need!}

%\thanks是放在\author的大括号里面~~  否则会在第一页出现空白页
\author{\IEEEauthorblockN{Xue Li,
Yuanzhi Cheng, \it{Member, IEEE}}
\thanks{
X. Li is with the School of Computer Science and Technology, Harbin Institute of Technology, Harbin, Heilongjiang, 150001, China (e\-mail: nefu\_education@126.com).

Y. Z. Cheng is with the School of Computer Science and Technology, Harbin Institute of Technology, Harbin, Heilongjiang, 150001, China, and also with the School of Information Science and Technology, Qingdao University of Science and Technology, Qingdao, Shandong 266061, China (e\-mail: yzcheng@hitwh.edu.cn).

%Corresponding author: Y. Z. Cheng. 

%Manuscript received Nov. 5, 2022. 
This work has been submitted to the IEEE for possible publication. Copyright may be transferred without notice, after which this version may no longer be accessible.
}}

%\markboth{Submitted Manuscript: IEEE Transactions on Pattern Analysis and Machine Intelligence}%
%{Xue Li \MakeLowercase{\textit{et al.}}: IEEE Transactions on Pattern Analysis and Machine Intelligence}



% for Transactions on Magnetics papers, we must declare the abstract and
% index terms PRIOR to the title within the \IEEEtitleabstractindextext
% IEEEtran command as these need to go into the title area created by
\IEEEtitleabstractindextext{%
\begin{abstract}
In designing and applying graph neural networks, we often fall into some optimization pitfalls, the most deceptive of which is that we can only build a deep model by solving over-smoothing. The fundamental reason is that we do not understand how graph neural networks work. Stress graph drawing can offer a unique viewpoint to message iteration in the graph, such as the root of the over-smoothing problem lies in the inability of graph models to maintain an ideal distance between nodes. We further elucidate the trigger conditions of over-smoothing and propose Stress Graph Neural Networks. By introducing the attractive and repulsive message passing from stress iteration, we show how to build a deep model without preventing over-smoothing, how to use repulsive information, and how to optimize the current message-passing scheme to approximate the full stress message propagation. By performing different tasks on 23 datasets, we verified the effectiveness of our attractive and repulsive models and the derived relationship between stress iteration and graph neural networks. We believe that stress graph drawing will be a popular resource for understanding and designing graph neural networks.
\end{abstract}

}

\maketitle

\IEEEdisplaynontitleabstractindextext

% \fi
\IEEEpeerreviewmaketitle

\section{Introduction}
\IEEEPARstart{A} graph $G(V, E)$ is an abstract structure that can model a relation $E$ over a set $V$ of entities in real life. Graph neural networks (GNNs) are becoming a standard means for analyzing graph-structured data. They generally follow a recursive message-passing scheme, where all nodes simultaneously compute a linear or nonlinear weighted combination of their neighbor messages. To date, many variants of message passing have been developed, and graph convolutional networks (GCNs) \cite{ref1} and graph attention networks (GATs) \cite{ref2} are the most widely used model architectures.

Starting with the success of transferring convolutions from Euclidean to non-Euclidean space, more and more researchers focus on integrating some popular techniques in deep learning, such as residual connection, dense connection, and attention mechanism, into GNNs. Constructing a deep model similar to ResNet \cite{ref3} to enhance the expressive power of GNNs is naturally becoming one of the most popular topics. However, deeply stacking GNN layers often results in significant drops in performance, even beyond just a few (2–4) layers. This performance deterioration is thought to be related to overfitting, vanishing gradients, and over-smoothing. The first two explanations can be ruled out in that the training and testing accuracies of GNNs are observed to degrade synchronously, and the shallow layer model is hard to cause a vanishing gradient effect \cite{ref4}. As for over-smoothing, we found that this term is wrongly used to cover all cases of building deep GNNs. As shown in Fig. 1, only when stacking GCN layers will the node representations of different classes quickly become indistinguishable. Deeply stacking linear GCN layers does not cause the performance deterioration presented in nonlinear iterations. Iterating linear GCN with random features (pubmed\_1000r) does not even cause any performance decay. Laplacian smoothing is often used to explain the mechanism of message passing in the literature, but we can not use it to figure out why the linear iteration can form such a stable node embedding (from the 30th to the 60th iteration).

\begin{figure}[!t]
\centering
\includegraphics[width=3.5in, height = 1.4in]{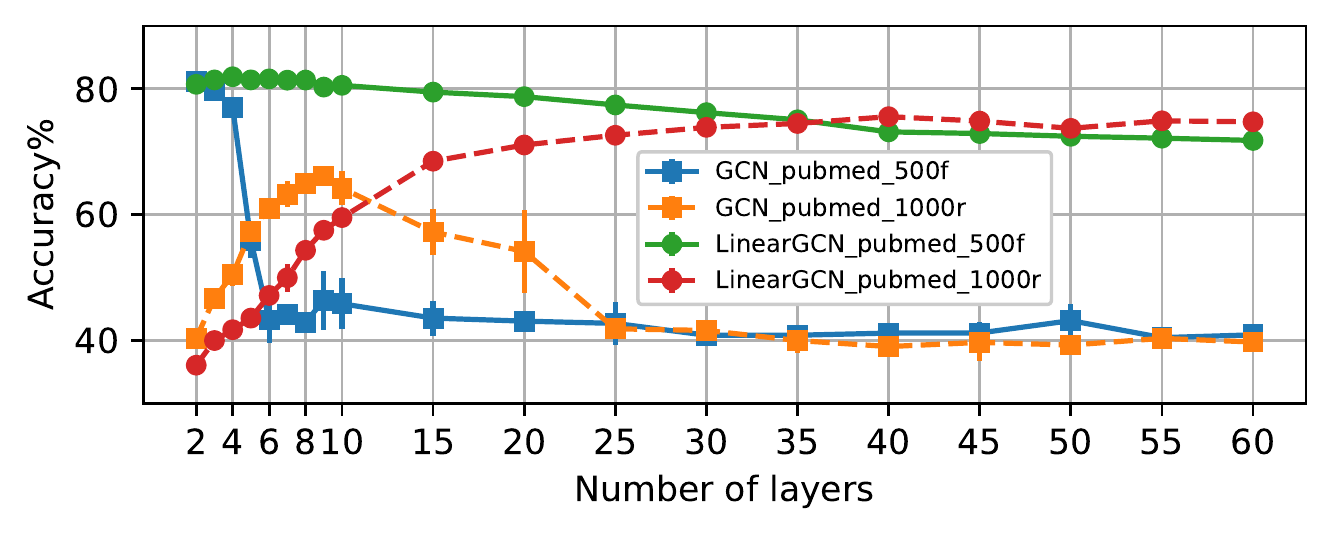}
% where an -eps-converted-to.pdf filename suffix will be assumed under latex, 
% and a .pdf suffix will be assumed for pdflatex; or what has been declared
% via \DeclareGraphicsExtensions.
\caption{Illustration of linear and nonlinear message propagations on the PubMed network with a 500-dimensional meaningful feature (pubmed\_500f) and a 1000-dimensional random (pubmed\_1000r) feature. The results are averaged over 5 runs and 100 epochs per run.}
%\label{fig_sim}
\end{figure}

Although some researchers have proposed several deep GNNs and claim that they have resolved the problem of over-smoothing, over-smoothing is inevitable in the current local message-passing scheme (especially for graphs with meaningful features). Based on the current definition of over-smoothing and explanations about message passing, the design of these deep GNNs can easily fall into the following optimization pitfalls:

\begin{itemize}
	\item [1.]
	 Redundant components: The model may be effective, but its structure is not the simplest design. 
	 \item [2.] 
	 Over-parameterization: Most of the model parameters can be pruned off.
	 \item [3.]
	 Error attribution: The model's validity is wrongly attributed to some unnecessary and unimportant components.
	 \end{itemize}

Deepening GNNs is still a controversial topic in the literature. In Section 4.2, we will redefine the concept of over-smoothing and systematically analyze the performance returns of deepening GNNs.

 Interestingly, stress graph drawing can shed some light on the answers to our confusion about GNNs. The core idea of stress visualization is to maintain an ideal distance (e.g., the shortest path distance) for every pair of nodes. Therefore, the original graph structure and the relative similarities of nodes can be well preserved in the final stress node embedding. Message iterations between connected nodes and between non-connected nodes help to ideally space all nodes. We refer to the message propagation between connected node pairs as attractive forces and the message propagation between non-neighboring node pairs as repulsive forces. The current message-passing scheme of GNNs is only a small part of stress message iteration. Over-smoothing can be completely prevented by stress iteration but at the cost of extremely high computational complexity ($O(n^2)$). By simplifying the stress iteration, we can build deep models with computational complexity similar to traditional GNNs. For the problem that deep GNNs may not be able to distinguish between structurally isomorphic nodes, we propose repulsive message passing, which propagates the distance-weighted messages between pairs of the target non-neighboring nodes. Moreover, we also propose the virtual pivot technique to integrate the repulsive component into current GNNs at the least cost.

Our main contribution is to answer the following questions:

\begin{itemize}
	\item [1.]
	 What is the geometric process of message passing?
	 \item [2.] 
	 What is over-smoothing?
	 \item [3.]
	 How to build a deep GNN?
	 \item [4.]
	 Is the model depth a resource or a burden on GNNs?
	 \item [5.]
	 How to optimize message passing according to stress graph drawing?
	 \end{itemize}

The remainder of this paper is organized as follows. Section 2 reviews some of the most common strategies for alleviating the problem of over-smoothing. Section 3 first presents two stress functions for computing graph layouts and then defines attractive and repulsive models based on stress iteration. In Section 4, we first conduct extensive experiments to explain what over-smoothing is and why we can build deep GNNs without preventing over-smoothing. Then, we verify the effectiveness of repulsive message passing and virtual pivots. Section 5 ends with a discussion and ideas for future research.

\section{Related Work}
 Inspired by the huge success of deep convolutional neural networks in computer vision, one might expect to increase the depth of GNNs to obtain more expressivity in describing the hierarchical information of graph datasets. However, it is known that stacking many layers and adding nonlinearity does not improve (or often worsen) the model's predictive performance. This effect, known as over-smoothing, has been regarded as the main bottleneck of GNN development. Recently, many endeavors have been made to alleviate the problem of over-smoothing. Some popular strategies for building deeper GNNs are listed below:
 
 \textbf{Residual Connections}: The earliest attempt at building deep GNNs can be traced back to 2017. Borrowing the concept from ResNet \cite{ref3}, Kipf and Welling \cite{ref1} used residual connection to connect the hidden layers of GCN (i.e., ResGCN). Unexpectedly, their experiments show that the residual connection alone is not sufficient to extend GCN to a deeper model. Chen et al. \cite{ref5} (2020) replaced the above residual connection with initial residual connection used in APPNP \cite{ref6} and further introduced the identity mapping technique from ResNet into GCN. The proposed model, GCNII, can be built to 64 layers. Its propagation rule is defined as follows:
\begin{small}
\begin{equation}
H^{l+1}=\sigma(((1-\alpha_l)PH^{(l)}+\alpha_lH^{(0)})\ ((1-\beta_l)I_n+\beta_lW^{(l)})))
\end{equation}
\end{small}where $\alpha_l$ and $\beta_l$ are two hyperparameters, $\sigma$ is an activation function, $P$ is the GCN aggregator, $H$ is the layer representation, and $W$ is the layer weight. The first term in the above equation is initial residual connection, which relieves over-smoothing significantly. The second term is identity mapping, which ensures that the deeper layer achieves at least the same performance as the shallow layer. The authors claimed that applying identity mapping and initial residual connection simultaneously ensures that the accuracy increases with the model depths. 

 \textbf{Jump/Dense Connections}: Xu et al. \cite{ref7} (2018) observed that the intermediate layer representations are also informative and proposed to combine different aggregations at the last layer selectively. The resulting architecture, JK-Net, is analogous to DenseNets \cite{ref8}. JK-Net achieves its best result with 6 layers on citation networks, and the authors concluded that global, together with local information, would help boost performance. Liu et al. \cite{ref9} (2020) attributed the performance deterioration of deep GNNs to the entanglement of layer weights and message iteration. By decoupling layer weights from message propagation and selectively combining the information from local and global neighborhoods, they proposed DAGNN, deepening GNNs up to 200 layers. Its architecture is defined as the following:
\begin{equation}
\left\{ 
\begin{array}{l}
Z = MLP(H^{(0)})\\ 
H_l = P^lZ, l=1,2,3,..,k\\ 
                 
H = \text{stack}(Z,H_1, ..., H_k) \\
S = \sigma(H \cdot t) \\

\end{array}
\right.
\end{equation} where $t$ is a trainable projection vector. Transformation, propagation, and dense connection adjustment constitute the method of DAGNN.

\textbf{Normalizations}: Different from above approaches, Zhao and Akoglu \cite{ref10} (2020) argued to keep the global pairwise node distance constant. Deep iterations would thus shrink intra-class distances while keeping unlinked node pairs relatively apart. The proposed normalization layer, PairNorm, works on each layer representation to slow down over-smoothing but does not necessarily increase the model performance. Zhou et al. \cite{ref11} (2020) developed a differentiable group normalization scheme (GroupNorm), which normalizes node features group-wise via a learnable soft cluster assignment:
\begin{equation}
S = \text{softmax}(HW)
\end{equation}where $W\in R^{F\times T}$ denotes a trainable weight matrix mapping each node into one of $T$ clusters. Normalization is then performed group-wise via:
\begin{equation}
H^{(l+1)} = H^{(l)} + \lambda\sum_{i=1}^{T}BatchNorm(S[:,i] \odot H^{(l)})
\end{equation}where $\lambda$ is a hyperparameter. Each learned group is normalized independently to increase the intra-class smoothness and to keep inter-class pairs farther off.

 \textbf{Graph Augmentations}: Feng et al. \cite{ref12} (2020) randomly removed some node features (i.e., DropNode) to generate $U$ perturbation matrices, which are then used to perform message propagation to produce $U$ augmented features. A consistency regularized loss is used to make the predictions of these $U$ augmented features tend to be the same. Their experiments on the Cora dataset show that a 10-layer GNN is still highly performant. Rong et al. \cite{ref13} (2020) randomly removed a certain rate of graph edges (i.e., DropEdge) to reduce the speed for the mixture of neighborhood features. The removal of edges in DropEdge is dynamic and layer-wise:
\begin{equation}
H^{(l+1)} = \sigma(\vartheta(A\odot B)H^{(l)}W^{(l)})
\end{equation}where $A$ is the adjacency matrix, $B$ is a binary random mask, and $\vartheta (A)=I_n+D^{-\frac{1}{2}}AD^{-\frac{1}{2}}$.

\textbf{Positional Awareness}: Although we can deepen GNNs by leveraging above techniques, You et al. \cite{ref14} (2019) found that arbitrarily deep GNNs still can not distinguish structurally isomorphic nodes. That is because most existing GNNs encode the local structure rather than the global node position into the embeddings. Consequently, nodes residing in similar subgraphs tend to be embedded to the same vector. To empower GNNs with the awareness of node positions, You et al. proposed the position-aware graph neural network (P-GNN). P-GNN randomly selects a certain number of nodes within the graph as anchor sets. Nodes learn their representations via aggregating the nearest anchor's features from different anchor sets. All features are weighted by the shortest path distance. Thus, any pair of embeddings can (approximately) recover their theoretical distance in the graph. This is the first work to address the need for encoding position information.

\textbf{Virtual Node}: Adding a virtual node to the graph is a well-known trick to improve the performance of graph classification. The original idea for this technique is to compute a graph embedding in parallel with the node embeddings. When connecting all real nodes to a virtual node in one direction, only the virtual node could receive information from all real nodes without affecting the message propagation of real nodes \cite{ref15}. When the virtual node is bidirectionally connected to all existing real nodes, the virtual node gets updated through message passing simultaneously as the real nodes do \cite{ref16}. The bidirectional setting is observed to alleviate over-smoothing to some extent \cite{ref17}. Although the virtual node is a well-known trick, its mechanism has never been theoretically investigated nor fully understood. It is worth further exploring the influence of one or more virtual nodes on building GNNs.

Almost all above optimization algorithms can slow down the model performance degradation, while only DAGNN and GCNII yield noticeable returns when increasing the depth. The dense connection of DAGNN and the identity mapping of GCNII, however, are not the essential components for alleviating over-smoothing. DAGNN and GCNII are redundant and can be further simplified. In fact, over-smoothing is natural and inevitable for the current message-passing scheme. The existing so-called deep GNN is just an imbalanced combination of shallow iterations (in dominance) and deep iterations. The virtual links used in PGNN and virtual nodes are a step in the right direction of completely preventing over-smoothing, which will be presented below. Moreover, the relationship among local message passing, PGNN, and virtual nodes will be systematically presented for the first time.

\section{Stress Graph Neural Networks}
\subsection{Preliminaries}

The earliest attempt at shortening the squared edge lengths in defining a nice layout can be traced back to the early 1960s, often referring to minimizing the following cost function.
\begin{equation}
\text{H}(p)=\sum_{<i,j>\in E}||p_i - p_j||^2
\end{equation}where node $i$ is placed at point $p_i$, and $E$ is the existing edges in the current graph. The partial derivative of $H(p)$ with respect to $p_i$ is
\begin{equation}
\frac{\partial H(p)}{\partial p_i}= 2\sum_{j\in N_i}(p_i-p_j)
\end{equation} where $N_i$ is the neighbor of node $i$. Equating Eq. (7) to zero and isolating the location of node $i$ gives
\begin{equation}
p_i = \frac{\sum_{j\in N_i}p_j}{\text{deg}(i)}
\end{equation}

This new derivation suggests that placing each node at the centroid of its neighbors could minimize $H(p)$. However, this solution is unwanted because nothing prevents all nodes from collapsing at the same location (equivalent to over-smoothing in GNNs). A simple but effective way of preventing nodes from getting too close to each other is to specify a uniform distance for every non-neighboring pair, as the following.
\begin{equation}
\text{G}(p)=\sum_{i\ne j \in V}(||p_i-p_j||-1)^2
\end{equation}

This cost function uniformly spaces all nodes and is completely independent of the target graph structure. Seeking a balance between $H(p)$ and $G(p)$ is an intuitive solution to graph drawing. The following equation is structured as a sum of two cost functions, one with target distances equal to 0 and the other with target distances equal to 1. This function is called the binary stress (B-Stress) function \cite{ref18}.
\begin{equation}
\text{B}(p)=\sum_{<i,j>\in E}||p_i-p_j||^2+\theta\sum_{i\ne j \in V}(||p_i-p_j||-1)^2
\end{equation} where $\theta$ is the balance coefficient. The first term emphasizes the local structure by making the graph layout of connected nodes more compact, while the second term strives to distribute all nodes evenly within a circle. To get some feeling, we provide results for three graphs in the first column of Fig. 2. The structure of these graphs is well presented. In this paper, unless stated otherwise, graph visualization data is from the SuiteSparse Matrix Collection \cite{ref19}.

To draw more details of the original graph structure, an ideal distance $d_{ij}$ for every pair of nodes $i$ and $j$ should be given. The associated cost function, full stress (F-Stress) \cite{ref20, ref21}, generally sets $d_{ij}$ as the length of the shortest path between nodes $i$ and $j$. We present the full stress function in an m-dimensional layout—$P\in R^{n\times m}$ as follows,
\begin{equation}
\text{Stress}(P)=\sum_{i<j}w_{ij}(||P_i-P_j||-d_{ij})^2
\end{equation} where $w_{ij}$ in this paper is taken as $d_{ij}^{-2}$. We can further decompose Eq. (11) to obtain
\begin{equation}
\begin{split}
\text{Stress}(P) = \sum_{<i,j>\in E}w_{ij}(||P_i-P_j||-d_{ij})^2 +\\ \sum_{<i,j>\in { V \choose 2 }\backslash E}w_{ij}(||P_i-P_j||-d_{ij})^2
\end{split}
\end{equation}

The first term is called attractive forces, which tend to shorten edges and maintain the compactness of connected nodes. Repulsive forces in the second term keep all nodes $V$ well separated. As shown in the second column of Fig. 2, the global layouts of the drawing produced by F-Stress faithfully and nicely represent the graph's structure. 

The F-Stress function can be further written in the form of the following iterative process (on axis $a$):
\begin{equation}
p_i = \frac{\sum_{j\ne i}w_{ij}(p_j+d_{ij}(\hat{p}_i-\hat{p}_j)\text{inv}(||P(t)_i-P(t)_j||))}{\sum_{j\ne i}w_{ij}}
\end{equation} where $\hat{p} = P(t)^{(a)}$ denotes the current coordinate, and $p= P(t+1)^{(a)}$ represents the new coordinate, which is calculated via the majorization optimization process (see [21] for more details). $\text{inv}(x) = 1/x$ when $x\ne 0$ and 0 otherwise.

This iteration ensures that node $i$ will be at distance $d_{ij}$ from $j$ in the final graph layout. Therefore, over-smoothing never happens. By extracting the attractive forces from Eq. (13) as follows, we can find that the current message propagation ($\sum_{i,j\in E}w_{ij}p_j$) in GNNs is only a small part of stress iteration.
\begin{small}
\begin{equation}
\begin{split}
p_{i}=\frac{\sum_{i,j\in E}(w_{ij}p_j+ w_{ij}d_{ij}(\hat{p}_i-\hat{p}_j)\text{inv}(||P(t)_i-P(t)_j||))}{\sum_{j\ne i}w_{ij}}
\end{split}
\end{equation}
\end{small}

\begin{figure}[h]
	
 \begin{minipage}{0.48\linewidth}
 	\vspace{3pt}
 	\centerline{\includegraphics[width=1.3in]{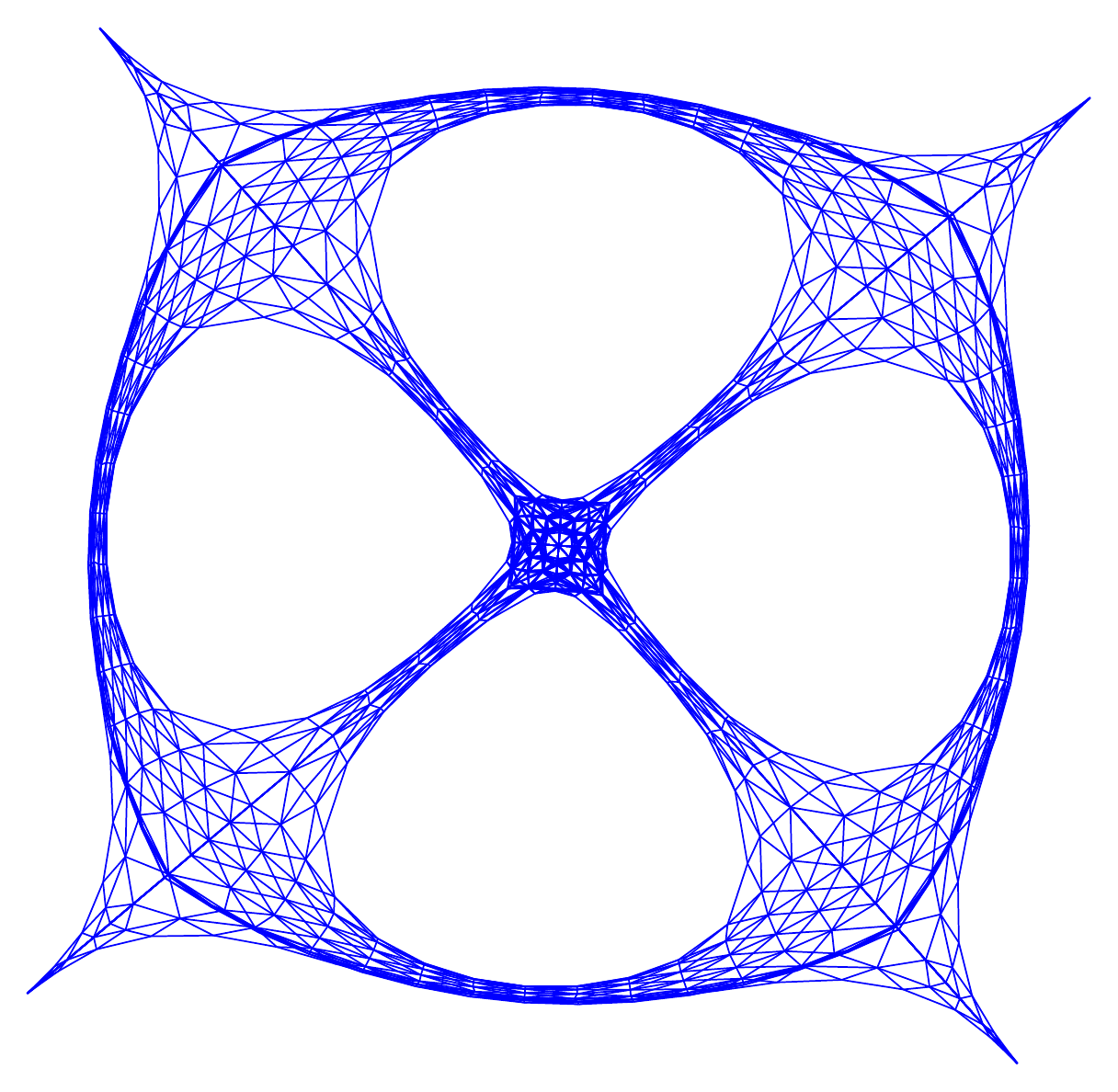}}
 	\vspace{3pt}
 	\centerline{\includegraphics[width=1.3in]{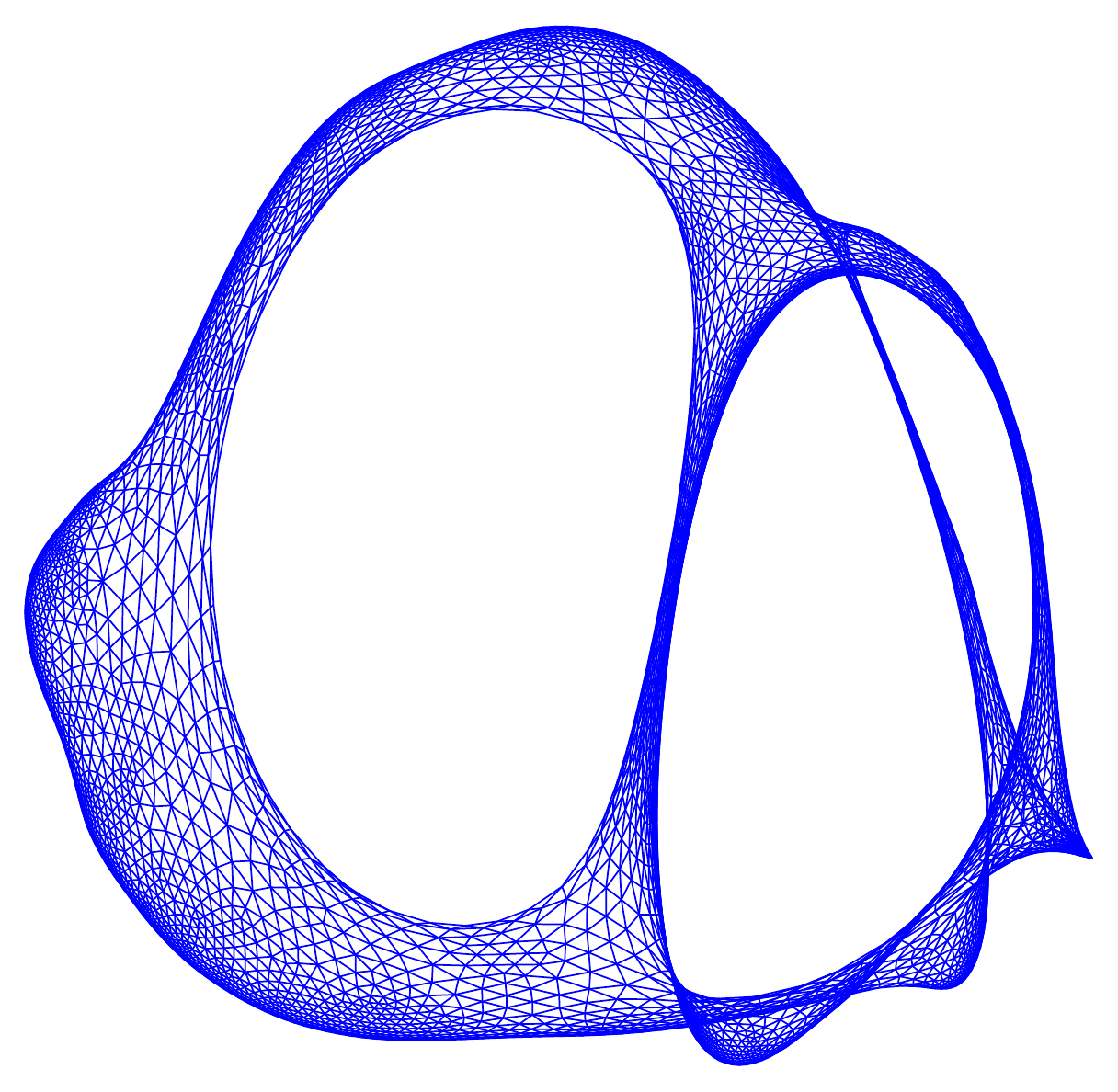}}
 	\vspace{3pt}
 	\centerline{\includegraphics[width=1.3in]{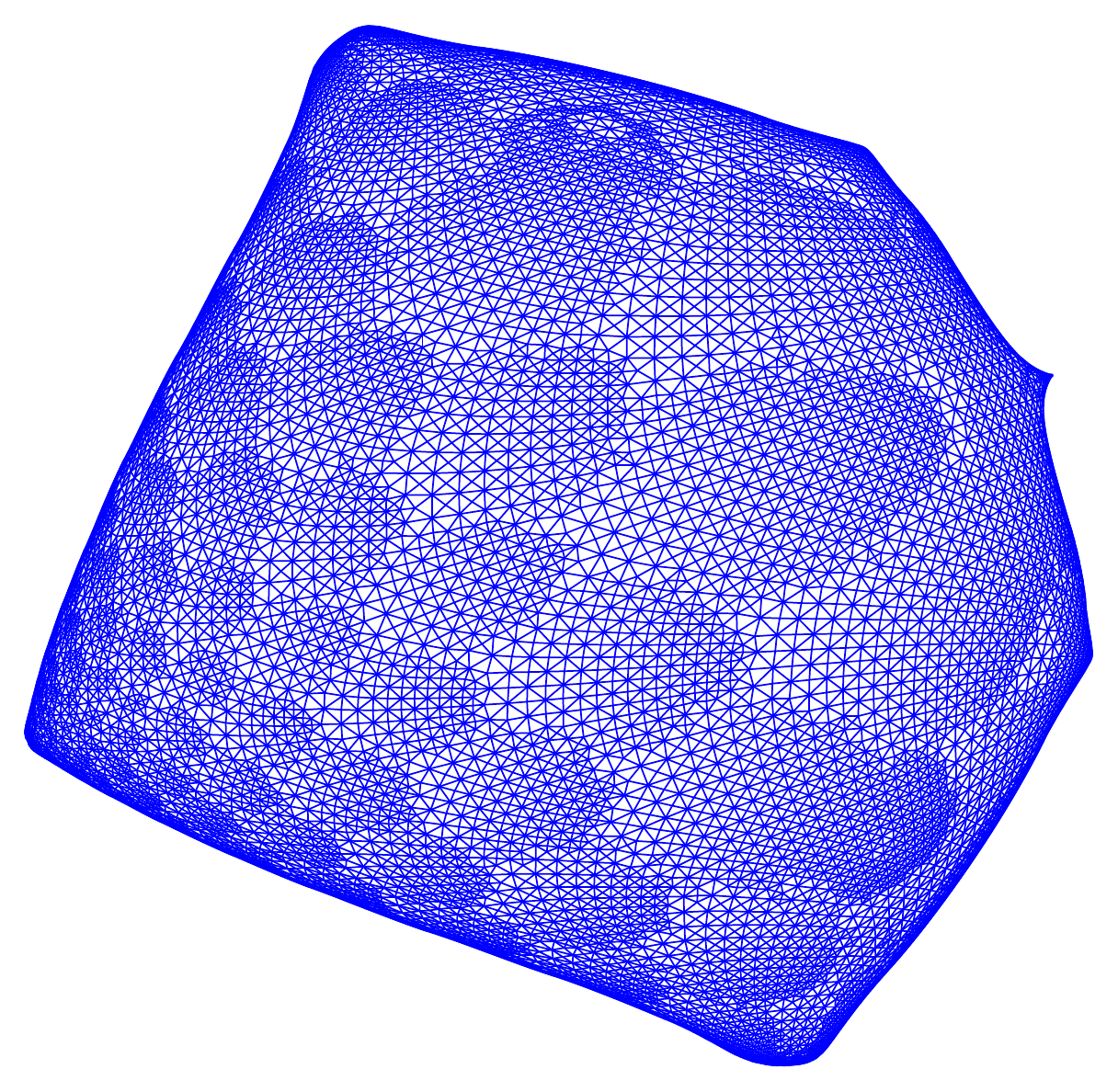}}
 	\vspace{3pt}
 	\centerline{Binary Stress}
 \end{minipage}
 \begin{minipage}{0.48\linewidth}
 	\vspace{3pt}
 	\centerline{\includegraphics[width=1.3in]{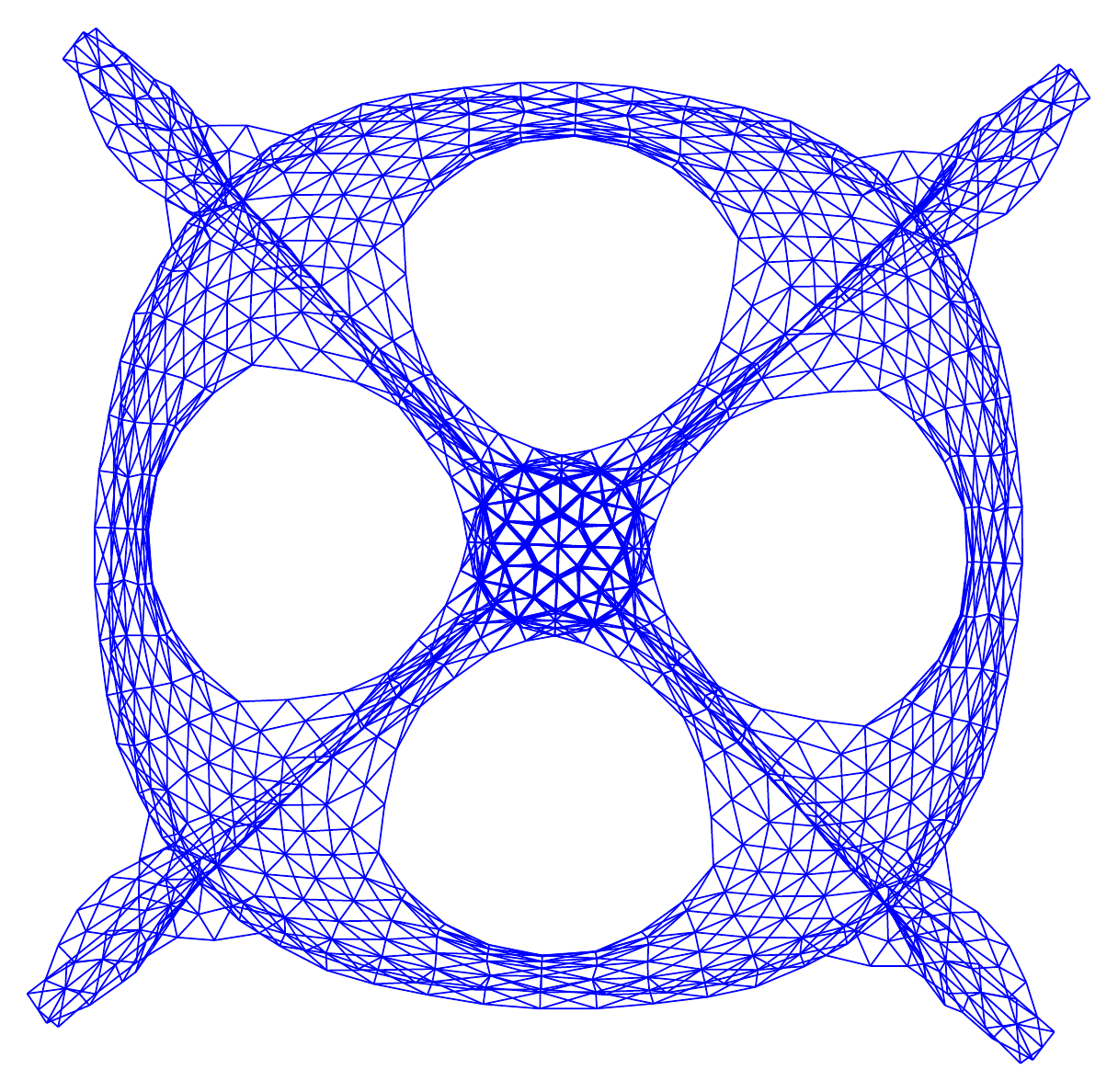}}
 	\vspace{3pt}
 	\centerline{\includegraphics[width=1.3in]{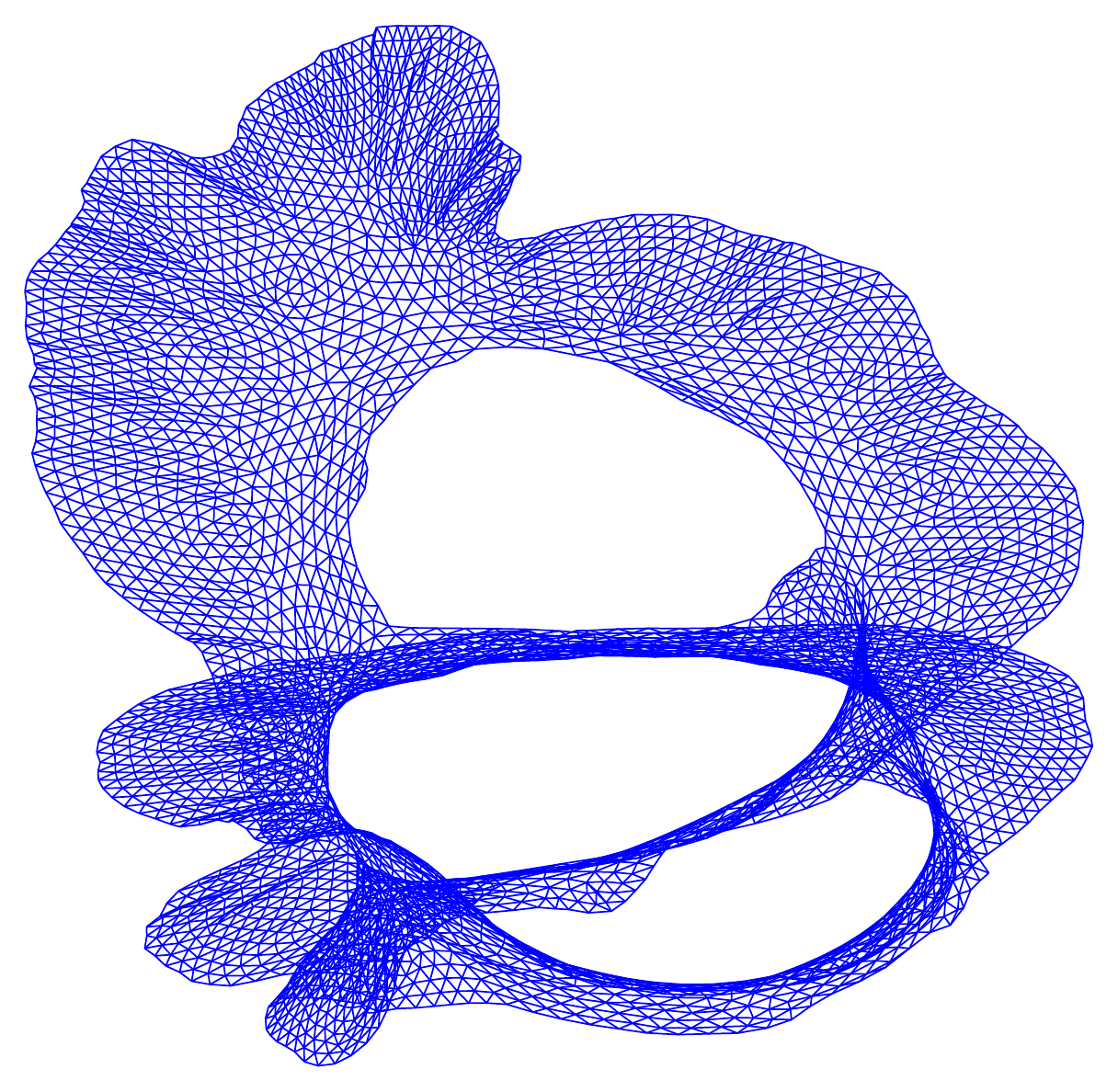}}
 	\vspace{3pt}
 	\centerline{\includegraphics[width=1.3in]{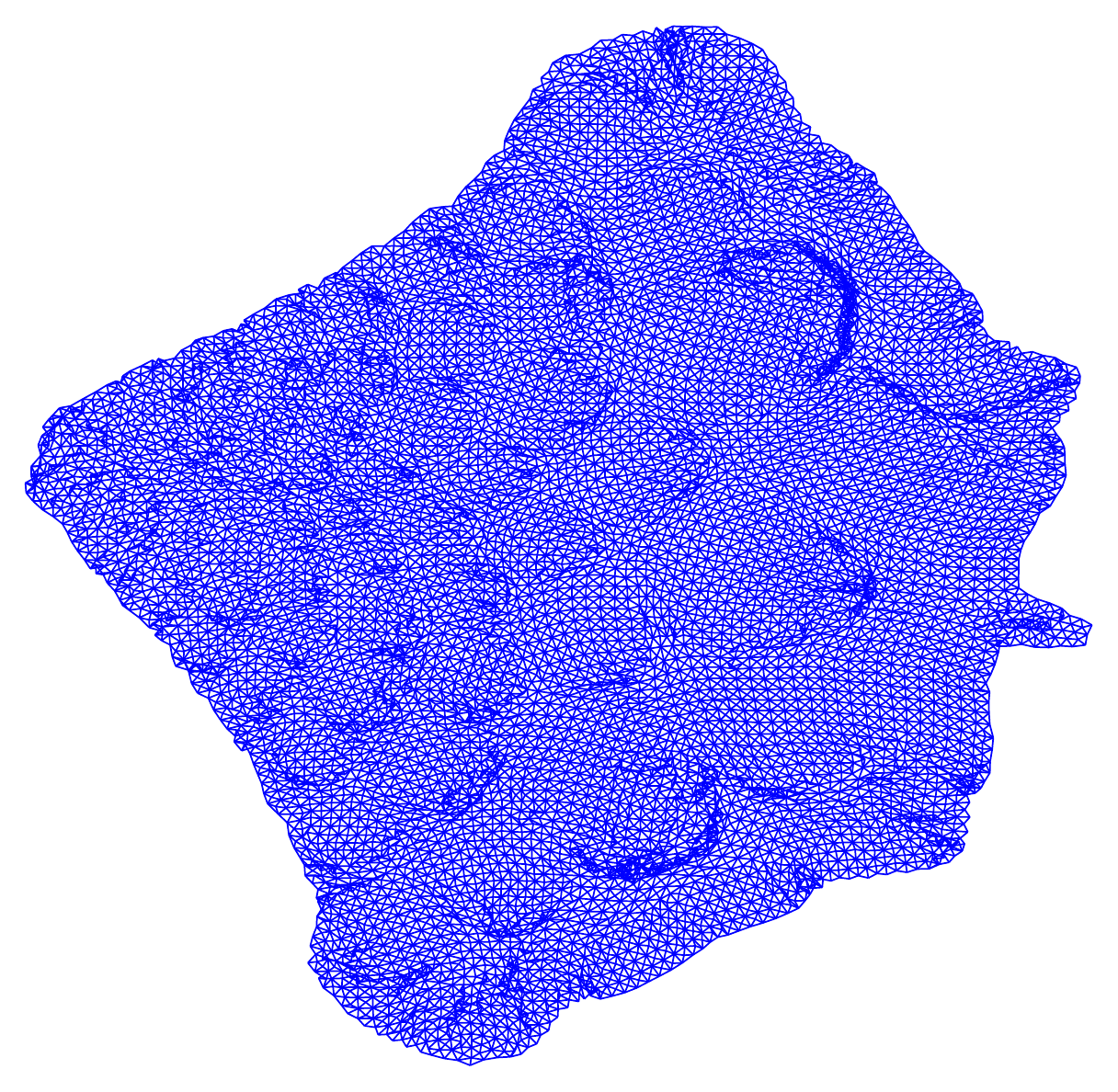}}
 	\vspace{3pt}
 	\centerline{Full Stress}
\end{minipage}

\caption{Examples of binary stress and full stress on the graphs, from top to bottom, dwt\_1005 (1,005 vertices), airfoil1 (4,253), and crack (10,240).}
\label{Fig.2}

\end{figure}

In the following subsections, we will present the theories and models of stress-based graph neural networks.

\subsection{Attractive Models}

Following the idea of stress graph drawing, we seek to maintain a simplified and efficient distance for each connected pair to deepen GNNs under the framework of the current message passing. We explored the following two scenarios: linear and nonlinear message propagations.

\vspace{0.3cm}
\noindent \textit{3.2.1 Linear Attractive Models}
\vspace{0.1cm}

The first model is concerned with the following local stress function.
\begin{equation}
X_i=\frac{\sum_{j\in N_i}M_{ij}(X_j+d_{ij}(X_i-X_j)\text{inv}(||X_i-X_j||))}{\sum_{j\in N_i}M_{ij}}
\end{equation} where $X_i$ is the input graph feature of node $i$, $M$ is the edge weight matrix (e.g., Laplacian matrix) and $d_{ij}$ is set to $M_{ij}^{-\beta},\beta =1,2...$. 

Since the shallow GNNs generally achieve state-of-the-art results, when building deep models, we keep the output of the best shallow layer $z$ as the spacing distance:
\begin{small}
\begin{equation}
X_i=\frac{\sum_{j\in N_i}M_{ij}(X_j+d_{ij}(X_i^{[z]}-X_j^{[z]})\text{inv}(||X_i^{[z]}-X_j^{[z]}||))}{\sum_{j\in N_i}M_{ij}}
\end{equation}
\end{small}

By setting $M$ to $I + D^{-1}A$ and $I + D^{-\frac{1}{2}}AD^{-\frac{1}{2}}$, we can propose StressDA and StressDAD, respectively. We denote DA model as $(I+D^{-1}A)^kX$ and DAD model as $(I + D^{-\frac{1}{2}}AD^{-\frac{1}{2}})^kX$. Since the repulsive force is absent in these models, all nodes still have the extra translation and rotation degrees of freedom. We can alleviate over-smoothing, but we can not prevent it. 

Although over-smoothing is inevitable in the local message-passing scheme, we can build GNNs into a deep model. By simplifying the distance in the above local stress, we can obtain
\begin{equation}
X_i = \sum_{j\in N_i}M_{ij}((1-\alpha)X_j+\alpha d_{ij}X_{0_i})
\end{equation} where $X_0$, the input graph feature, is used as a kind of distance. Just like the attraction-repulsion balance strategy in B-Stress, $\alpha$ is used to control the proportion of the initial distance $X_0$ that can be maintained between every pair of connected nodes. We always take $\alpha =0.1$, which seems to achieve the best performance in most cases. $d_{ij}$ can be further set to $1/\sum_{j\in N_i}M_{ij}$, then we obtain
\begin{equation}
X_i = \sum_{j\in N_i}M_{ij}(1-\alpha)X_j+\alpha X_{0_i}
\end{equation}

By setting $M$ as Laplacian matrices, we can propose S-StressDA (simplified StressDA) and S-StressDAD (or called APPNP [6]), respectively. Residual connection here is given the attribute of distance. These models can be deepened infinitely, and in Section 4.3, we will explain why we can successfully build deep GNNs without resolving over-smoothing.

\vspace{0.3cm}
\noindent \textit{3.2.2 Nonlinear Attractive Models}
\vspace{0.1cm}

The distance in above simplified stress models is manually set. How should we modify these models to learn the node distance adaptively? Identity mapping and dense connection emphasized in previous deep GNNs, as mentioned in Section 2, are not our answers. We simply transform the input graph feature $X_{in}$ as follows and keep the remaining structure unchanged.
\begin{equation}
\begin{array}{l}
X_0=\sigma(X_{in}W_{in})	\\
X_i = \sum_{j\in N_i}M_{ij}(1-\alpha)X_j+\alpha X_{0_i}
\end{array}
\end{equation} where $\sigma$ is the activation function and $W_{in}$ is the learnable weight. Stacking all node vectors in the matrix yields
\begin{equation}
X^{(k)} = (1-\alpha)MX^{(k-1)}+\alpha X_0,\quad X^{(0)}=X_0
\end{equation}

By setting $M$ to $I + D^{-\frac{1}{2}}AD^{-\frac{1}{2}}$, we can propose StressGCN.

Since there are no repulsive forces in StressGCN, the distance between non-neighboring nodes is not in our control. We can not use the pure distance maintenance theory in stress iteration to explain the global node-position relationship. More noteworthy is the side effect of specifying only the local distance—the cumulative iteration. The main difference between StressGCN and StressDAD is that the former carefully designs the proportion of shallow and deep layers (see Section 4.3.1 for more details). DAGNN and GCNII also try to select the intermediate iteration results adaptively, but their model structures are redundant. The simplest and most effective structure we can extract from DAGNN and GCNII is StressGCN. A detailed comparative analysis of these three models is provided in Section 4.3.

\begin{figure}[h]

\begin{minipage}{0.48\linewidth}
 	\vspace{3pt}
 	\centerline{\includegraphics[width=1.3in]{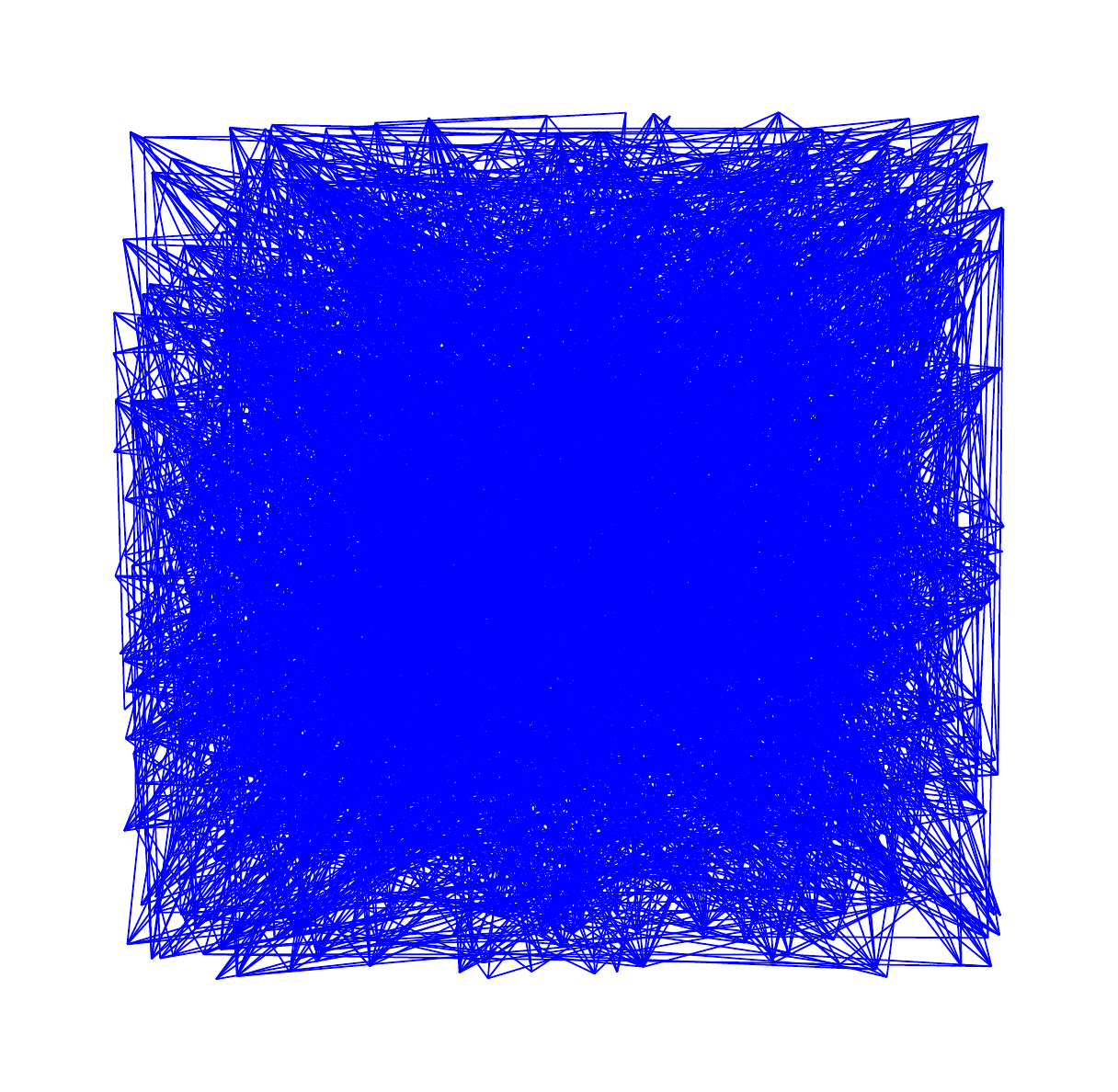}}
 	\vspace{3pt}
 	\centerline{(a)}
 	\centerline{\includegraphics[width=1.3in]{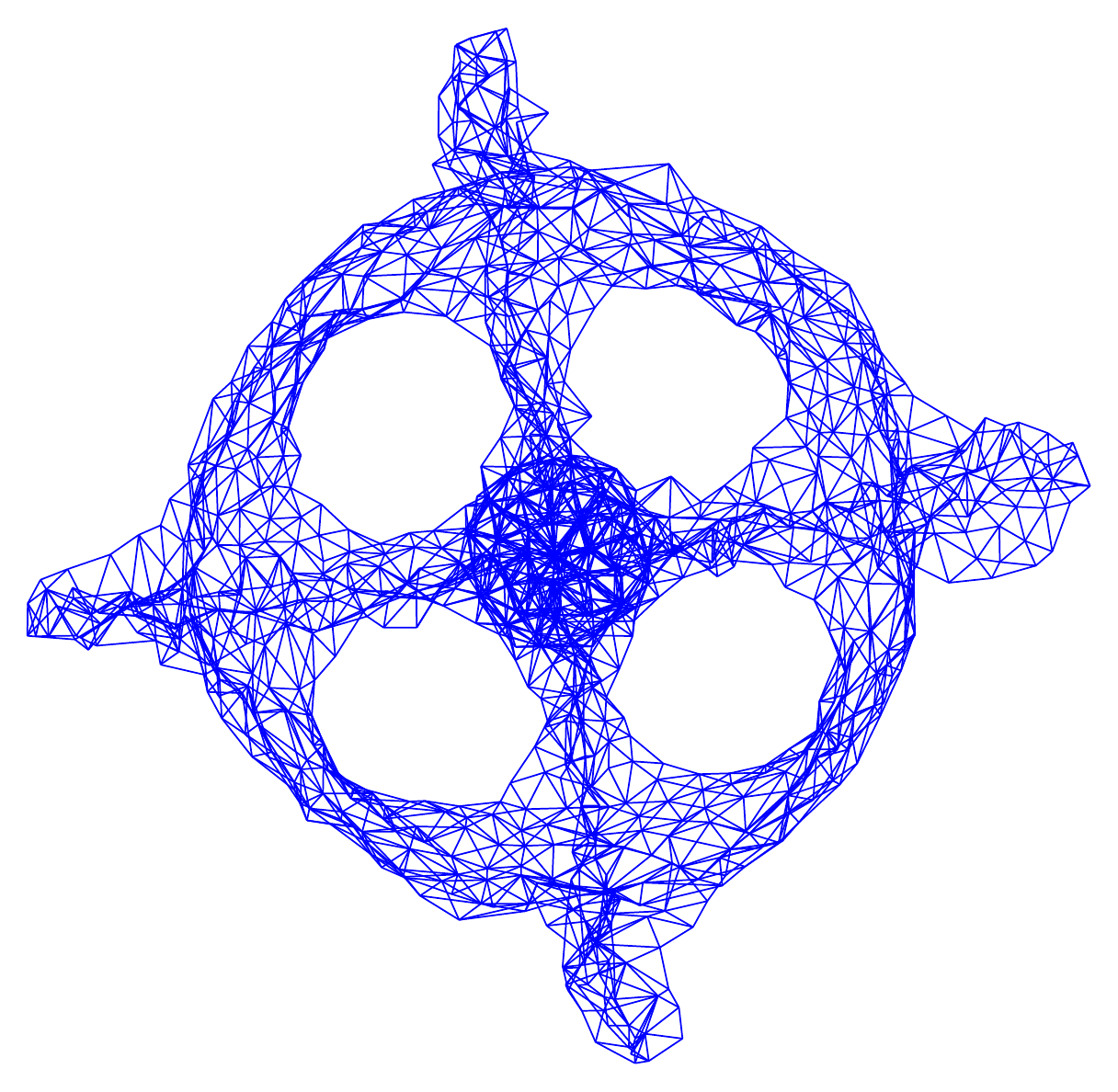}}
 	\vspace{3pt}
 	\centerline{(c)}

 \end{minipage}
 \begin{minipage}{0.48\linewidth}
 	\vspace{3pt}
 	\centerline{\includegraphics[width=1.3in]{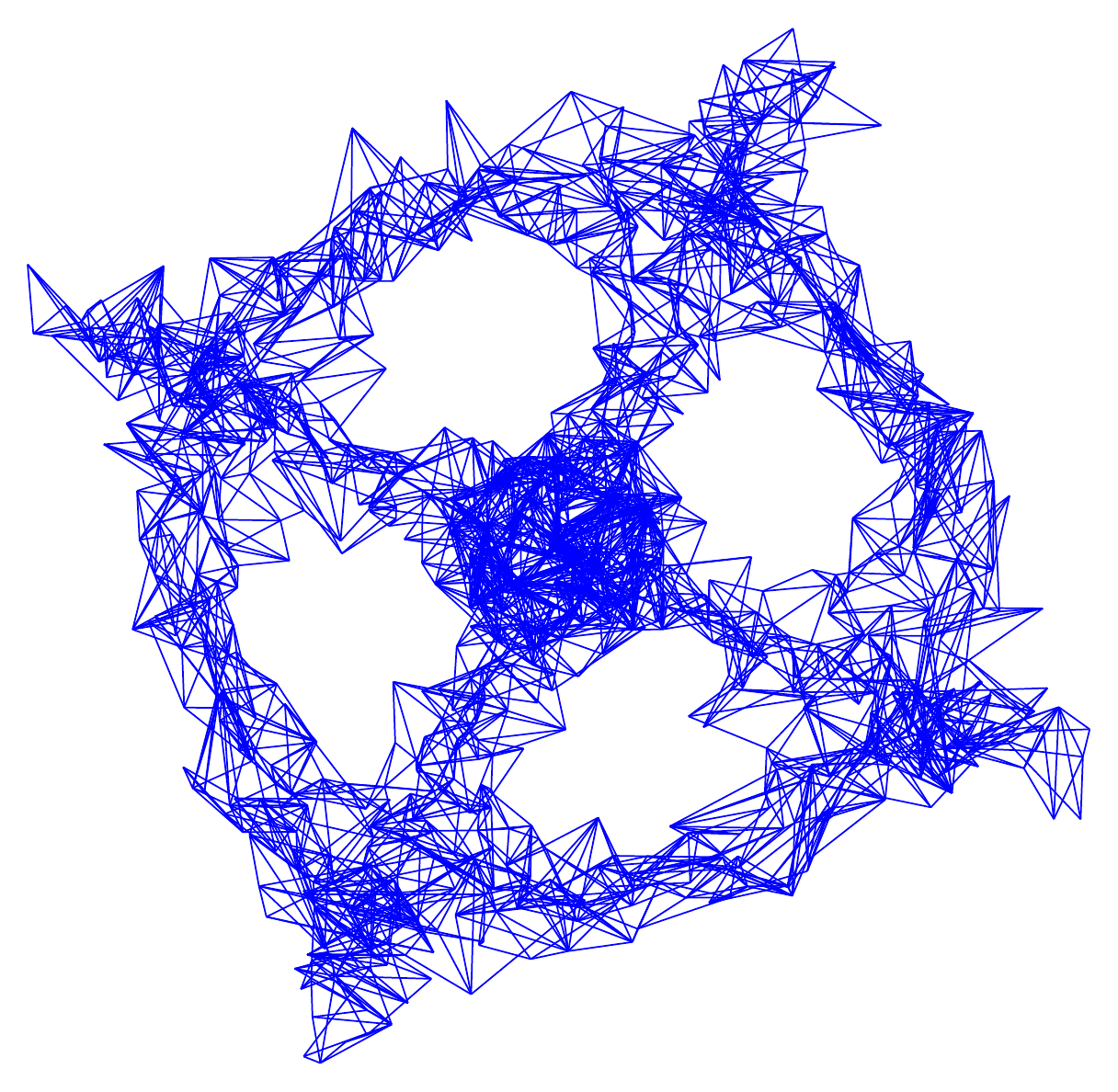}}
 	\vspace{3pt}
 	\centerline{(b)}
 	\centerline{\includegraphics[width=1.3in]{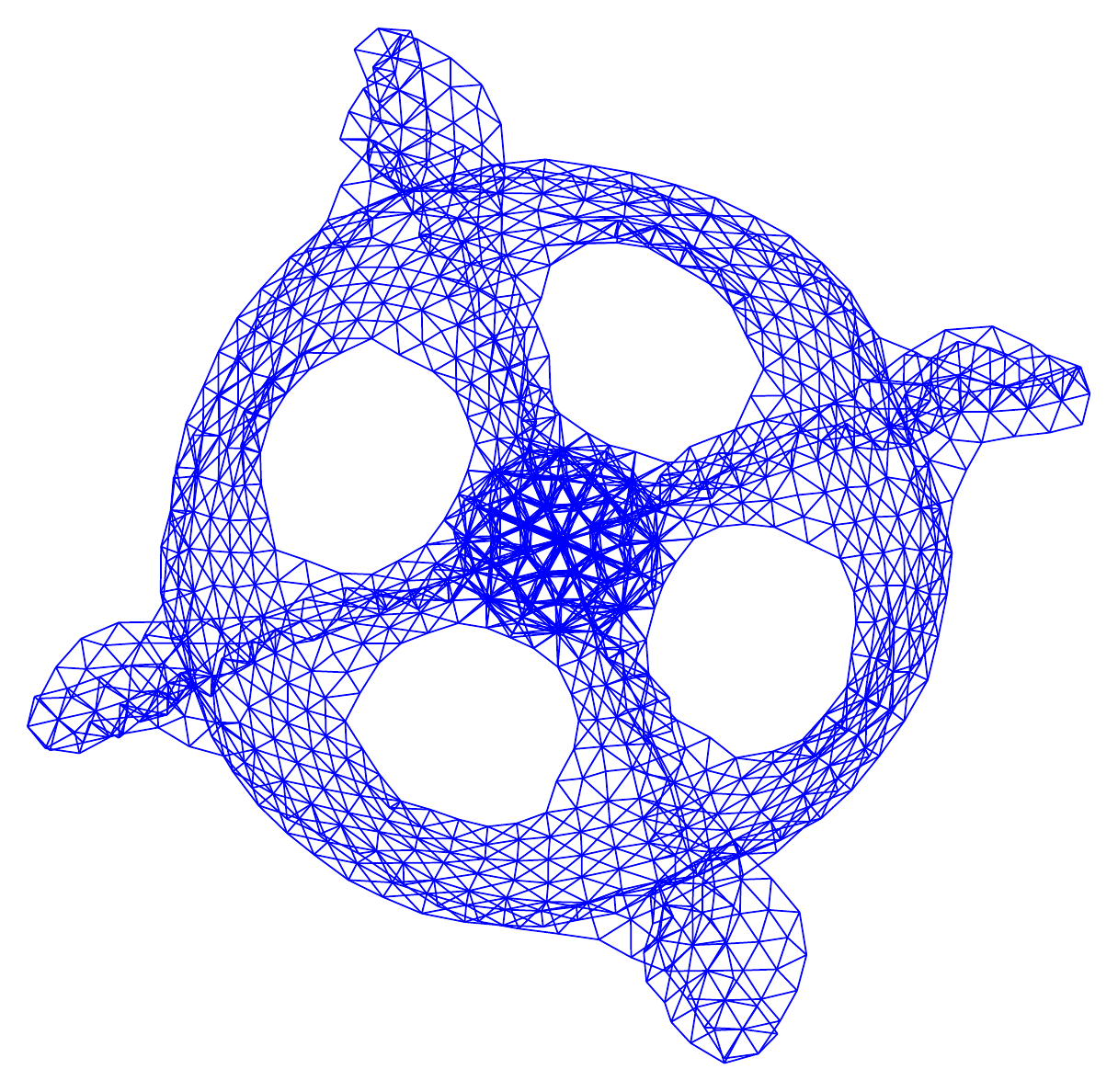}}
 	\vspace{3pt}
 	\centerline{(d)}
\end{minipage}
\caption{Example run of sparse stress on the graph dwt\_1005 with random initialization (a) and intermediate iteration results (b,c,d).}
\label{Fig.3}
\end{figure}

\subsection{Repulsive Models}

The full stress method in graph drawing has many advantages, such as minimizing the edge crossings and maximizing the layout symmetry. However, it is not practical to calculate all-pairs shortest distances for large graphs anyway. In this subsection, we will explore how to simplify repulsion and how to apply it in GNNs better.

\vspace{0.3cm}
\noindent \textit{3.3.1 Sparse Repulsion Models}
\vspace{0.1cm}

Ortmann et al. \cite{ref22} proposed the sparse stress function to reduce the high computations in the full stress method. They restrict the stress computation of each node $i$ to a set of representative nodes $\mathcal{P}$, from now on called pivots. The simplified stress function has the following form:
\begin{equation}
\begin{split}
\text{Sparse}(X)= \sum_{<i,j>\in E}w_{ij}(||X_i-X_j||-d_{ij})^2+\\ \sum_{i\in V}\sum_{p \in \mathcal{P}\backslash N_i}w_{ip}(||X_i-X_p||-d_{ip})^2
\end{split}
\end{equation} where $d_{ip}$ is the distance between node $i\in V$ and pivot $p\in V$. By selecting 200 pivots (discussed later) for sparse stress, we present the formation of the final layout of the graph dwt\_1005 in Fig. 3. The original graph structure and the relative similarities of the nodes are still well maintained.

Instead of directly using sparse stress to design GNNs, we focus on the information exchange between each node $i$ and all pivots as follows, where $M_{ip}=(d_{ip}+1)^{-1}$.
\begin{equation}
X_i= \sum_{p \in \mathcal{P}}M_{ip}X_p,\quad i\in V
\end{equation}

\newcommand{\INDSTATE}[1][1]{\STATE\hspace{#1\algorithmicindent}}
\begin{algorithm} 
	\caption{HDE-Pivot \cite{ref23}, Proposed in 2002} 
	\label{alg1} 
	\begin{algorithmic}[1]
		\STATE \textbf{Pivot Selection} $(G(V={1,…,n},E), m)$ 
		\STATE \% This function finds $m$ pivots
		\STATE Choose node $p_1$ randomly from $V$
		\STATE dist[$1,…,n$]$\leftarrow\infty$\quad \% a distance recording list
		\FOR{ $i = 1$ to $m$}
		\STATE \% Compute the shortest path distance
		\STATE $\mathcal{D}_{p_i*}\leftarrow BFS(G(V,E),p_i)$
		\FOR{every $j\in V$}
		\STATE dist[j]$\leftarrow$min\{dist[j], $\mathcal{D}_{p_ij}$\}
		\ENDFOR
		\STATE \% Choose next pivot
		\STATE $p_{i+1}\leftarrow$ arg max$_{j\in V}$\{dist[$j$]\}
		\ENDFOR
		\RETURN $p_1,p_2,...,p_m$
	\end{algorithmic} 
\end{algorithm}

Unlike local message passing, every node receives a distance-weighted message from the well-designed pivots instead of from neighbors. We refer to this kind of message propagation as repulsive message passing. The final output can be further combined with the input graph feature via concatenation or product. The resulting model is called the sparse repulsion graph neural network (SR-GNN).

The position-aware graph neural network (P-GNN) is the most closely related work. In P-GNN, the representative nodes are called anchors, similar to our pivots. PGNN selects sets of anchor nodes and aggregates information from these anchors. The main difference between SR-GNN and P-GNN is the sampling strategies of representative nodes. In Algorithm 1 and Algorithm 2, we compare the pivot and the anchor selection methods.

There is no substantial difference between pivots and anchors, and we use the two terms to distinguish SR-GNN from P-GNN. The shortest distance can be computed using breadth-first-search (BFS) or any other method (e.g., Dijkstra’s algorithm). Inspired by P-GNN, we design the following two variants of SR-GNN:

\begin{itemize}
	\item [1.]
	 \textbf{SR-GNN-F}, using the truncated shortest path distance (e.g., 2-hop distance).
	 \item [2.] 
	 \textbf{SR-GNN-E}, using the exact shortest path distance.
	 \end{itemize}
	 
P-GNN samples anchors in each forward pass, while SR-GNN samples pivots only once. In Section 4.4, we will compare the static and dynamic modes of the pivot sampling. 

Although You et al. proposed to recover the shortest path distance between nodes in the node embedding, P-GNN actually can not maintain the node distance consistent with graph-theoretic target distances. It just integrates the distance information into the weights of virtual edges.

Moreover, to find the globally optimal position of each node $i$, the sparse stress function also sets $w_{ip}$ in Eq. (21) as adaptive weights. By recording the closest nodes to each pivot $p$, we can define a pivot control region $\mathcal{R}(p)$. $w_{ip}$ is then set to $\psi/d_{ip}^2$, where $\psi=|j\in\mathcal{R}(p):d_{jp}\le d_{ip}/2|$ (see \cite{ref22} for more details). By setting $M_{ip}$ in Eq. (22) to $\psi/d_{ip}$, we can propose SR-GNN-R. In subsequent experiments, SR-GNN-E, SR-GNN-F, and SR-GNN-R will be thoroughly compared with P-GNN-E and P-GNN-F.

\vspace{0.3cm}
\noindent \textit{3.3.2 Virtual Pivots}
\vspace{0.1cm}

\begin{algorithm} 
	\caption{P-GNN-Anchor \cite{ref14}, Proposed in 2019} 
	\label{alg2}
	\begin{algorithmic}[1]
		\STATE \textbf{Anchor Selection} $(G(V={1,…,n},E), t)$ 
		\STATE \% This function finds $t$ anchor sets
		\STATE Set anchor size = $\text{int}(n/(2^{j+1}))$
		\INDSTATE[5.7] $,j=0,1,...,log_n-1$
		\STATE Each size is chosen for $clog_n$ times
		\INDSTATE[6],c=0.5,1,...
		\STATE Randomly sample $t=clog_n^2$ anchor sets
		\INDSTATE[5.7] $,S=\{S_1,S_2,...S_t\}$
		\FOR{ $i = 1$ to $t$}
		\STATE \% Compute the shortest path distance
		\STATE $\mathcal{D}_{S_i*}\leftarrow BFS(G(V,E),S_i)$
		\STATE Choose the closest anchor for each node
		\ENDFOR
		\RETURN ($v$, $s_1$, $s_2$, ..., $s_t$), $v\in V, s_i\in S_i$
	\end{algorithmic} 
\end{algorithm}

\renewcommand{\thetable}{\arabic{table}}
\begin{table}
\caption{Dataset statistics for different tasks.}
\label{table-1}
\centering

\setlength{\tabcolsep}{1.33mm}{
\begin{tabular}{c|llll}
\hline
{\color[HTML]{333333} \textbf{Task}}                                                                                     & {\color[HTML]{333333} \textbf{Dataset}} & {\color[HTML]{333333} \textbf{\#Nodes}} & {\color[HTML]{333333} \textbf{\#Edges}} & {\color[HTML]{333333} \textbf{\#Features}} \\ \hline
{\color[HTML]{333333} }                                                                                                  & {\color[HTML]{333333} Cora}             & {\color[HTML]{333333} 2,708}            & {\color[HTML]{333333} 5,429}            & {\color[HTML]{333333} 1,433}               \\
{\color[HTML]{333333} }                                                                                                  & {\color[HTML]{333333} CiteSeer}         & {\color[HTML]{333333} 3,327}            & {\color[HTML]{333333} 4,732}            & {\color[HTML]{333333} 3,703}               \\
{\color[HTML]{333333} }                                                                                                  & {\color[HTML]{333333} PubMed}           & {\color[HTML]{333333} 19,717}           & {\color[HTML]{333333} 44,338}           & {\color[HTML]{333333} 500}                 \\
\multirow{-4}{*}{{\color[HTML]{333333} \textbf{\begin{tabular}[c]{@{}c@{}}Node \\ Classification\end{tabular}}}}         & {\color[HTML]{333333} Ogbn-arxiv}       & {\color[HTML]{333333} 169,343}          & {\color[HTML]{333333} 1,166,243}        & {\color[HTML]{333333} 128}                 \\ \hline
{\color[HTML]{333333} }                                                                                                  & {\color[HTML]{333333} Grid-1}           & {\color[HTML]{333333} 400}              & {\color[HTML]{333333} 760}              & {\color[HTML]{333333} 1}                   \\
{\color[HTML]{333333} }                                                                                                  & {\color[HTML]{333333} Com-1}    & {\color[HTML]{333333} 400}              & {\color[HTML]{333333} 3800}             & {\color[HTML]{333333} 1}                   \\
{\color[HTML]{333333} }                                                                                                  & {\color[HTML]{333333} Grid-400}         & {\color[HTML]{333333} 400}              & {\color[HTML]{333333} 760}              & {\color[HTML]{333333} 400}                 \\
\multirow{-4}{*}{{\color[HTML]{333333} \textbf{\begin{tabular}[c]{@{}c@{}}Link\\ Prediction\end{tabular}}}}              & {\color[HTML]{333333} Com-400}  & {\color[HTML]{333333} 400}              & {\color[HTML]{333333} 3800}             & {\color[HTML]{333333} 400}                 \\ \hline
{\color[HTML]{333333} }                                                                                                  & {\color[HTML]{333333} Cornell}          & {\color[HTML]{333333} 183}              & {\color[HTML]{333333} 295}              & {\color[HTML]{333333} 1703}                \\
{\color[HTML]{333333} }                                                                                                  & {\color[HTML]{333333} Texas}            & {\color[HTML]{333333} 183}              & {\color[HTML]{333333} 309}              & {\color[HTML]{333333} 1703}                \\
{\color[HTML]{333333} }                                                                                                  & {\color[HTML]{333333} Wisconsin}        & {\color[HTML]{333333} 251}              & {\color[HTML]{333333} 499}              & {\color[HTML]{333333} 1703}                \\
\multirow{-4}{*}{{\color[HTML]{333333} \textbf{\begin{tabular}[c]{@{}c@{}}Pairwise Node\\ Classification\end{tabular}}}} & {\color[HTML]{333333} Email}            & {\color[HTML]{333333} 920/7}            & {\color[HTML]{333333} 7201/7}           & {\color[HTML]{333333} 1}                   \\ \hline
{\color[HTML]{333333} }                                                                                                  & {\color[HTML]{333333} MSRC\_9}          & {\color[HTML]{333333} 41/221}           & {\color[HTML]{333333} 196/221}          & {\color[HTML]{333333} 10}                  \\
{\color[HTML]{333333} }                                                                                                  & {\color[HTML]{333333} MSRC\_21C}        & {\color[HTML]{333333} 40/209}           & {\color[HTML]{333333} 193/209}          & {\color[HTML]{333333} 22}                  \\
{\color[HTML]{333333} }                                                                                                  & {\color[HTML]{333333} Cuneiform}        & {\color[HTML]{333333} 21/267}           & {\color[HTML]{333333} 90/267}           & {\color[HTML]{333333} 3}                   \\
{\color[HTML]{333333} }                                                                                                  & {\color[HTML]{333333} Ogbg-molbace}     & {\color[HTML]{333333} 34/1513}          & {\color[HTML]{333333} 74/1513}          & {\color[HTML]{333333} 9}                   \\
\multirow{-5}{*}{{\color[HTML]{333333} \textbf{\begin{tabular}[c]{@{}c@{}}Graph\\ Classification\end{tabular}}}}         & {\color[HTML]{333333} Ogbg-molbbbp}     & {\color[HTML]{333333} 24/2309}          & {\color[HTML]{333333} 52/2309}          & {\color[HTML]{333333} 9}                   \\ \hline
\end{tabular}
}
\end{table}

The B-Stress function simplifies the connected and non-neighboring pair distances to 0 and 1, respectively. The underlying graph of B-Stress contains a binary node distance (e.g., 1 and 2). How can we construct such a graph? Adding a virtual node to connect all nodes is an interesting answer. Although it is not something new for GNNs, the design of virtual nodes has never been theoretically investigated nor fully understood. The below pivot-based model may push our understanding of the virtual node.
\begin{equation}
\left\{ 
\begin{array}{l}
X_i = \sum_{j\in N_i}M_{ij}X_i+X_{p},\quad i\in \mathcal{R}(p)\\
X_p=\sum_{i\in \mathcal{R}(p)}w_{pi}X_{i}
\end{array}
\right.
\end{equation} where $\mathcal{R}(p)$ is the control region of pivot $p$, and $X_p$ is initialized to zero, and $w_{pi}$ can be set to 1 or $1/|\mathcal{R}(p)|$.

When we define only one pivot, this pivot is connected with all nodes, which is equivalent to the virtual node. When selecting multiple pivots using the pivot selection method presented in Algorithm 1, each node in the region $\mathcal{R}(p)$ is connected with pivot $p$. The whole process is more like the combination of B-Stress and sparse stress. By combing the virtual pivot technique with GCN, we can propose GCN-VP. In Section 4.4, we will explore the influence of virtual pivots on over-smoothing and graph visualization.

\section{Experiments}

\subsection{Experimental Setup}

\vspace{0.3cm}
\noindent \textit{4.1.1 Tasks and Datasets}
\vspace{0.1cm}

\textbf{Node Classification}: We answer questions, such as what is over-smoothing and how to build deep models, by experimenting on four citation networks: Cora, CiteSeer, PubMed \cite{ref24}, and ogbn-arxiv \cite{ref25}. The first three networks are PyTorch built-in data, and the last is from Open Graph Benchmark (OGB) leaderboard. We use the official train/validation/test splits for training.

\textbf{Link Prediction}: We use the following datasets to verify the effectiveness of sparse repulsive models:

\textbf{• Grid-1 \cite{ref14}}. A 2-dimensional grid graph consisting of 400 nodes and a 1-dimensional feature vector.

\textbf{• Com-1 \cite{ref14}}. A synthetic connected caveman graph [26] with 20 communities where each community has 20 nodes. 1\% of the graph edges are randomly rewired. The graph has a 1-dimensional feature vector.

\textbf{• Grid-400 and Com-400 \cite{ref14}}. The graph structures are the same as described above, but each node has a 400-dimensional feature vector.

\textbf{Pairwise Node Classification} : The task is to predict whether two nodes belong to the same community. We experiment on the following datasets to compare repulsive message passing with local message passing:

\textbf{• Web networks}. The nodes and edges of Cornell, Texas, and Wisconsin \cite{ref27} represent web pages and hyperlinks, respectively. Each node has a 1703-dimensional feature vector.

\textbf{• Email}. 7 real-world email communication networks from SNAP \cite{ref28}. Each graph has 6 communities and a 1-dimensional feature vector.

For link prediction and pairwise node classification tasks, we follow the setting used in the experiments of P-GNN \cite{ref14}. 80\% of existing links and an equal number of non-existing links are used for training, and the remaining two sets of 10\% of links are used as test and validation sets.

\textbf{Graph Classification}: We use the following datasets to verify the effectiveness of the virtual pivot models:

\textbf{• Vision networks}. MSRC\_9, MSRC\_21, and Cuneiform \cite{ref29} are image networks. e.g., Cuneiform represents the network of ancient wedge-shaped symbols. 

\textbf{• Biomolecular networks}. Ogbg-molbace and ogbg-molbbbp are OGB datasets \cite{ref25}, and each node has a 9-dimensional feature vector. The task is to predict whether a molecule has specific properties. e.g., whether a molecule can inhibit virus replication or not. We use the official train/validation/test splits for training.

Table 1 lists the summary statistics for each dataset. When multiple graphs are available, we present ``average nodes/num of graphs" and ``average edges/num of graphs" for the number of nodes and edges, respectively. In this paper, we report the test set performance when the best performance on the validation set is achieved, and we report results over 10 runs.

\vspace{0.3cm}
\noindent \textit{4.1.2 Baseline models}
\vspace{0.1cm}

We evaluate the performance of our stress graph neural networks by comparing them with several baselines. 1) State-of-the-art GNNs: GCN, SAGE \cite{ref30}, DAD (linear GCN), DA (linear SAGE). 2) Deep GNNs: ResGCN, Initial-ResGCN, GCN(group\_norm), GCN(drop\_edge), JKNetGCN, DAGNN, DeeperGCN \cite{ref31}, and GCNII. 3) Position-aware models: P-GNN and its variants. 4) Virtual node model: GCN-VN (denoted as GCN-VP1in this paper). All models are listed in Table 2.

\subsection{What is Over-smoothing?}

\vspace{0.3cm}
\noindent \textit{4.2.1 Three Stages of Message Iteration}
\vspace{0.1cm}

\begin{figure*}[!t]
 \centering
 \includegraphics[width=6.7in, height = 2.5in]{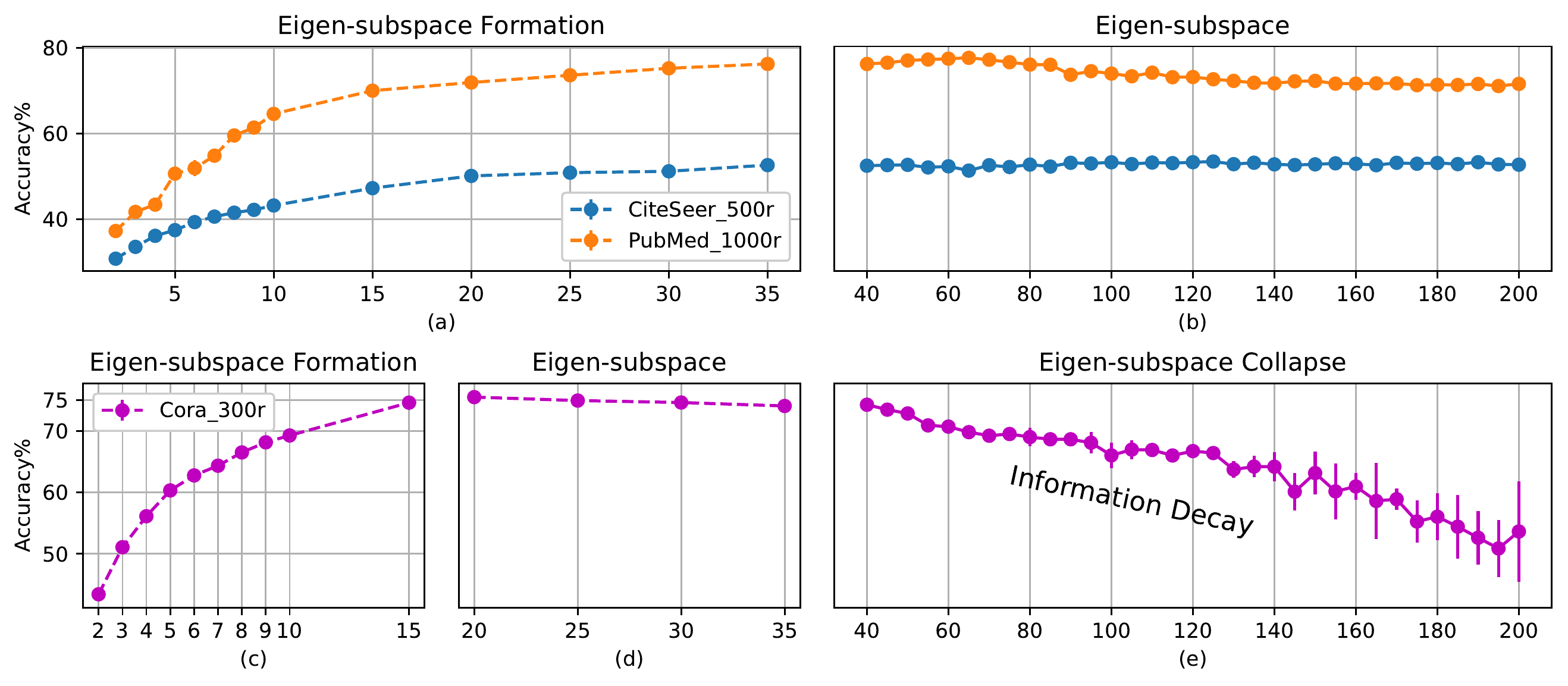}
% where an -eps-converted-to.pdf filename suffix will be assumed under latex, 
% and a .pdf suffix will be assumed for pdflatex; or what has been declared
% via \DeclareGraphicsExtensions.
 \caption{Illustration of the eigen-subspace of Laplacian aggregator-based models. We set $M=I + D^{-1}A$ and randomly generate 300, 500, and 1000 dimensional features for Cora, CiteSeer, and PubMed datasets denoted as Cora\_300r, CiteSeer\_500r, and PubMed\_1000r, respectively. The X-axis denotes the number of iterations. The results are averaged over 10 runs and 100 epochs per run.}
%\label{fig_sim}
\end{figure*}

\renewcommand{\thetable}{\arabic{table}}
\begin{table}
\caption{Stress GNNs and baseline models.}
\label{table-2}
\centering

\setlength{\tabcolsep}{0.72mm}{
\begin{tabular}{c|cc}
\hline
                                                                    & StressDA                     & StressDAD                   \\ \cline{2-3} 
\multirow{-2}{*}{\textbf{Linear Stress Models}}                     & S-StressDA                   & S-StressDAD                 \\ \hline
\textbf{Linear GNNs}                                                & DA                           & DAD                         \\ \hline
\rowcolor[HTML]{FFFF00} 
\textbf{Deep Stress Model}                                          & \multicolumn{2}{c}{\cellcolor[HTML]{FFFF00}StressGCN}      \\ \hline
\rowcolor[HTML]{FFFF00} 
\cellcolor[HTML]{FFFF00}                                            & GCN                          & ResGCN                      \\ \cline{2-3} 
\rowcolor[HTML]{FFFF00} 
\cellcolor[HTML]{FFFF00}                                            & \multicolumn{2}{c}{\cellcolor[HTML]{FFFF00}Initial-ResGCN} \\ \cline{2-3} 
\rowcolor[HTML]{FFFF00} 
\cellcolor[HTML]{FFFF00}                                            & GCN(group\_norm)             & GCN(drop\_edge)             \\ \cline{2-3} 
\rowcolor[HTML]{FFFF00} 
\cellcolor[HTML]{FFFF00}                                            & JKNetGCN                     & DAGNN                       \\ \cline{2-3} 
\rowcolor[HTML]{FFFF00} 
\multirow{-5}{*}{\cellcolor[HTML]{FFFF00}\textbf{Deep GNNs}}        & DeeperGCN                    & GCNII                       \\ \hline
\rowcolor[HTML]{22D2CA} 
\cellcolor[HTML]{22D2CA}                                            & SR-GNN-E                     & SR-GNN-F                    \\ \cline{2-3} 
\rowcolor[HTML]{22D2CA} 
\multirow{-2}{*}{\cellcolor[HTML]{22D2CA}\textbf{Sparse Repulsive Models}} & SR-GNN-R                     & SR-GNN-F(Dynamic)           \\ \hline
\rowcolor[HTML]{22D2CA} 
\textbf{Position-aware Models}                                      & P-GNN-E                      & P-GNN-F                     \\ \hline
\textbf{Virtual Pivot}                                              & \multicolumn{2}{c}{GCN-VP}                                 \\ \hline
\textbf{Virtual Node}                                              & GCN                          & GCN-VN                      \\ \hline
\end{tabular}
}
\end{table}

Let us begin by analyzing the process of linear message passing. The expression $M^kX$ is actually known as power iteration \cite{ref32}, an ancient method to compute eigenvectors. Assume that the matrix $M$ has n eigenvectors $x_1, x_2,...,x_n$ with corresponding eigenvalues of $\lambda_1,\lambda_2,...,\lambda_n$, in descending order. Then a nonzero random starting vector $v_0$ has
\begin{equation}
v_0 = c_1x_1+c_2x_2+...+c_nx_n,\quad c_i\ne0
\end{equation}

Multiplying both sides of this equation by $M$, we obtain
\begin{equation}
\begin{aligned}
Mv_0 &= c_1(Mx_1)+c_2(Mx_2)+...+c_n(Mx_n)\\
&=c_1(\lambda_1x_1)+c_2(\lambda_2x_2)+...+c_n(\lambda_nx_n)
\end{aligned}
\end{equation}

Repeated multiplication of both sides of this equation by $M$ gives
\begin{equation}
M^kv_0=c_1(\lambda_1^kx_1)+c_2(\lambda_2^kx_2)+...+c_n(\lambda_n^kx_n)
\end{equation}

 According to the power iteration clustering theory \cite{ref33, ref34}, the above iteration will go through two stages:
 
\begin{itemize}
	\item [1.]
	 \textbf{Eigen-subspace Formation}: With increasing $k$, the eigenvectors with small eigenvalues will shrink quickly, and the remaining eigenvectors are preserved to form an informative subspace where the clusters are well-separated.
	 \item [2.] 
	 \textbf{Dominant Eigenvector Formation}: When $k$ is large enough, the eigen-subspace will collapse and converge to the dominant eigenvector (the eigenvector of the largest-in-magnitude eigenvalue) of $M$.
	 \end{itemize}

\begin{figure}[!t]
 \centering
 \includegraphics[width=3.0in, height = 1.1in]{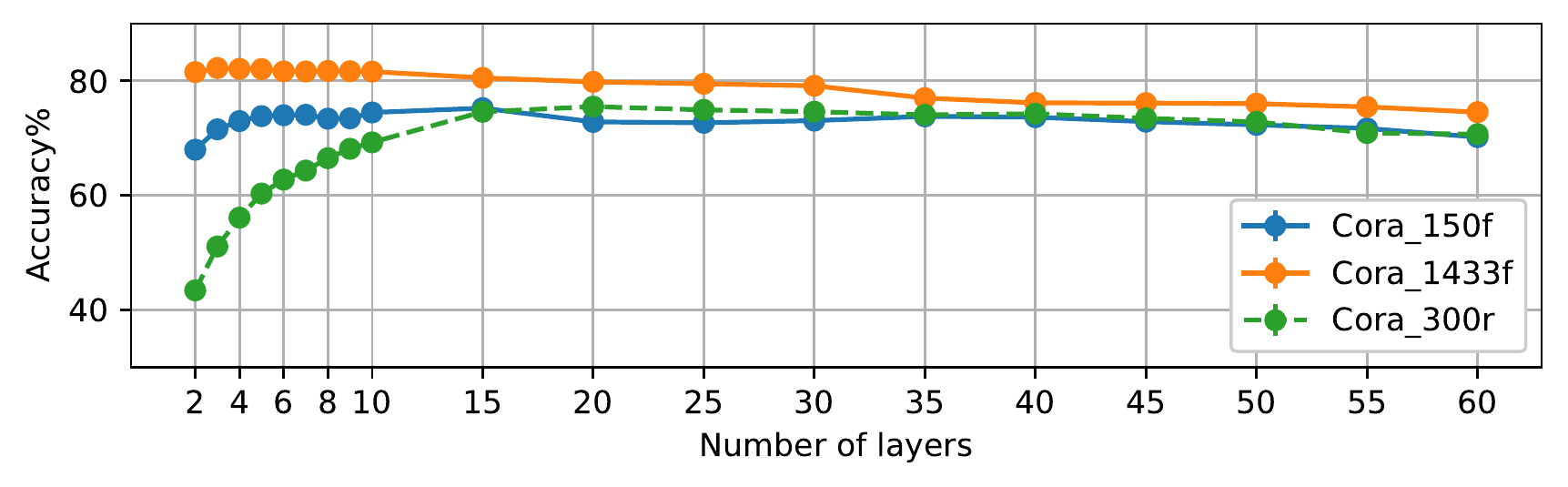}
% where an -eps-converted-to.pdf filename suffix will be assumed under latex, 
% and a .pdf suffix will be assumed for pdflatex; or what has been declared
% via \DeclareGraphicsExtensions.
 \caption{The feature information capacity test for varying starting iteration spaces.}
%\label{fig_sim}
\end{figure}

The eigen-subspace is crucial to understanding the status of message iteration. To obtain some intuitive insights into the relationship between eigen-subspaces and over-smoothing, we randomly generate the graph feature $X_r$ (e.g., Cora\_300r) and run $(I+D^{-1}A)^kX_r$ on citation benchmark networks. The formation, duration, and collapse of eigen-subspaces are presented in Fig. 4. The following four observations can be drawn from the evolution of eigen-subspaces.

\begin{itemize}
	\item [1.]
	 In Fig. 4 (a) and (c), the eigen-subspace contains more community information than the starting iteration space, so the eigen-subspace formation is an incremental iteration, and over-smoothing will not happen.
	 \item [2.] 
	 Fig. 4 (b) and (d) show the stability of the eigen-subspace, of which the quality and duration time depend on the data's exact nature. 
	 \item [3.] 
	 In Fig. 4 (e), the eigen-subspace contains more community information than the dominant eigenvector or fixed point. The dominant eigenvector of $(I+D^{-1}A)$ is $1_n=(1,...,1)$, that is, all nodes hold the same information. Over-smoothing in this stage is inevitable.
	 \item [4.]
	 A complete iteration process will go through the eigen-subspace's formation, duration, and collapse.
	 \end{itemize}

 \begin{figure*}[!t]
 \centering
 \includegraphics[width=6.8in, height = 2.4in]{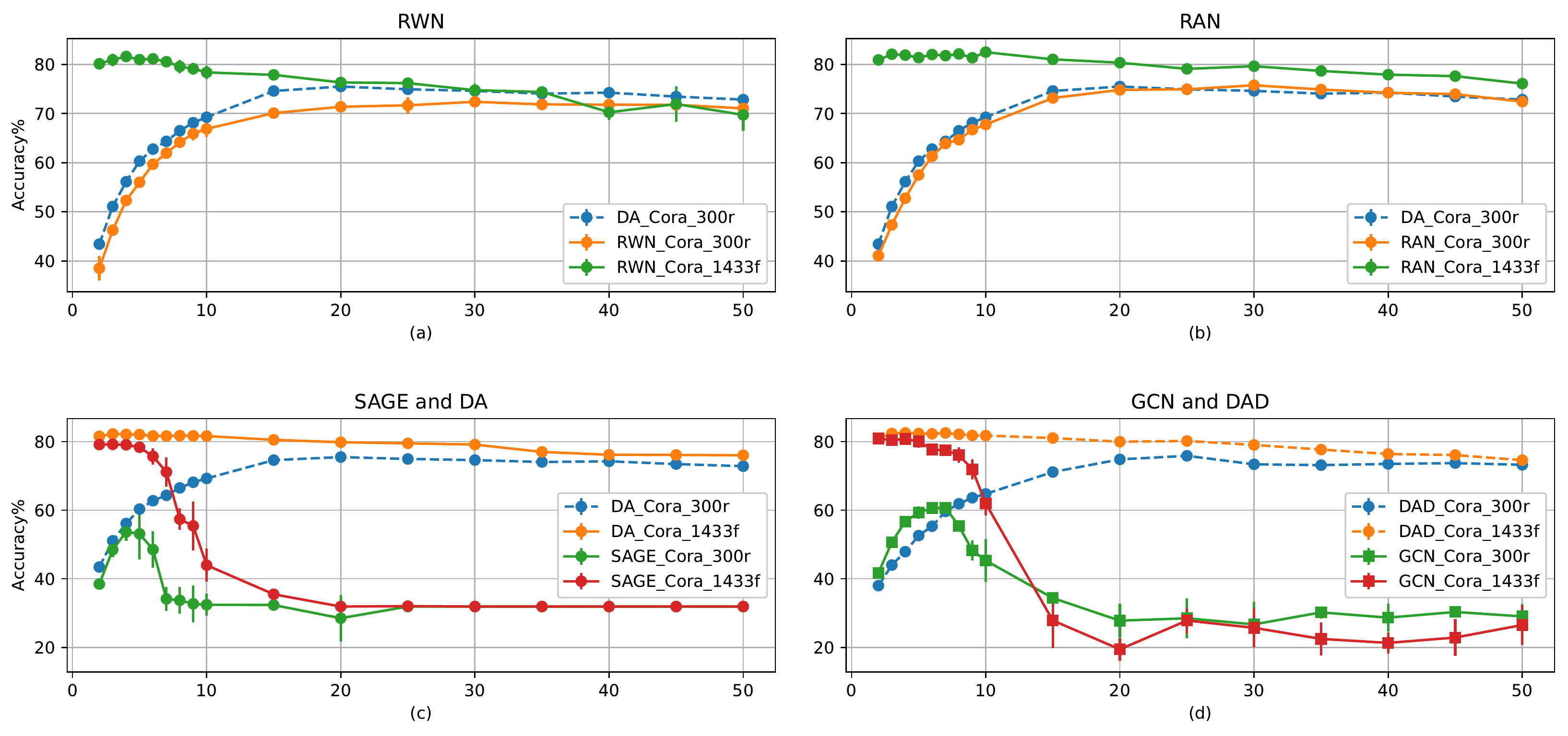}
% where an -eps-converted-to.pdf filename suffix will be assumed under latex, 
% and a .pdf suffix will be assumed for pdflatex; or what has been declared
% via \DeclareGraphicsExtensions.
 \caption{Illustration of irregular iterations and nonlinear iterations.}
%\label{fig_sim}
\end{figure*}

The next logical question is: what will happen if the information in the starting iteration space is greater than or equal to that of the eigen-subspace? We do this test on the Cora dataset. ``Richer information" corresponds to the Cora with a 1433-dimensional original feature (Cora\_1433f); ``equal information" is constructed by averaging the original feature into a 150-dimensional feature matrix (Cora\_150f). As shown in Fig. 5, only when the information in the starting iteration space is richer will the over-smoothing phenomenon happen. The trigger of over-smoothing actually depends on the input graph features' information capacity. For graphs with rich information in their features, simply stacking GNN layers inevitably results in information decay.

\begin{figure*}[!t]
\centering
\includegraphics[width=6.8in, height = 1.6in]{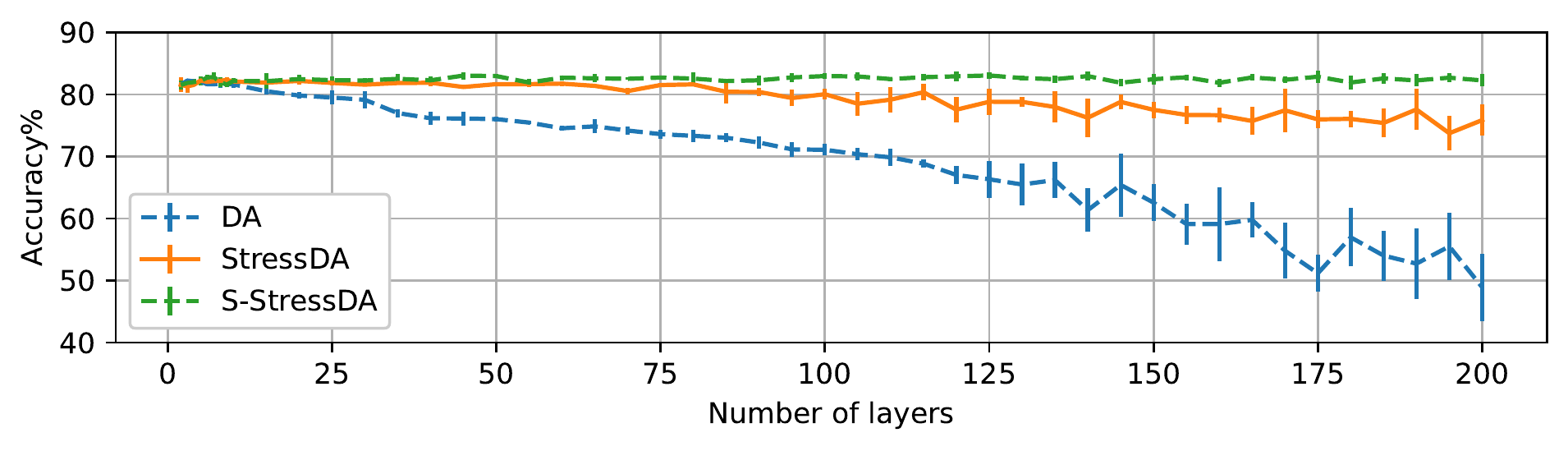}
% where an -eps-converted-to.pdf filename suffix will be assumed under latex, 
% and a .pdf suffix will be assumed for pdflatex; or what has been declared
% via \DeclareGraphicsExtensions.
\caption{Statistics of linear models at the 1000th iteration.}
%\label{fig_sim}
\end{figure*}

\vspace{0.3cm}
\noindent \textit{4.2.2 Irregular Iteration and Nonlinear Iteration}
\vspace{0.1cm}

Existing GNNs attach much importance to the design of the propagation weight matrix $M$, and the above power iteration also requires the matrix $M$ must be diagonalizable. However, in \cite{ref35}, we proved that the effectiveness of message passing is not uniquely tied to these well-designed propagation weights. Some “bad” edge weights could also achieve a similar effect. Hence, we introduce the following two irregular GNNs to study their iteration property.
\begin{itemize}
	\item [1.]
	 \textbf{RWN (Random Weight Networks)—$M_r^kX$}: The weights of $M_r$ are directly generated from the uniform or normal distribution (in absolute value), and each row is $L_2$-normalized.
	 \item [2.] 
	 \textbf{RAN (Random Attention Networks)—$M_{ran}^kX$}: A random attention vector (a $L_2$-normalized random vector) is used to calculate the weight of $M_{ran}$. The feature information is incorporated into the edge weights in a random way. 
	 \end{itemize}
	 
$M_r$ and $M_{ran}$ can be seen as a perturbation matrix of $I+D^{-1}A$ (see \cite{ref35} for more details).

 The corresponding GNN model of $I+D^{-1}A$ is denoted as DA. As shown in Fig. 6 (a) and (b), random weights and random attentions do not change the eigensystem too much. They can form a stable eigen-subspace through random feature iterations (RWN and RAN v.s. DA on Cora\_300r) and achieve a similar result to standard GNNs (e.g., SAGE and GCN) in shallow iterations on the Cora\_1433f dataset. 
 
 It is difficult to judge whether these aggregators are diagonalizable, but they indeed form an effective eigen-subspace. 
 Over-smoothing still originates from the information difference between the starting iteration space and the eigen-subspace.
 
 Thus far, the typical over-smoothing phenomenon is not discussed, that is, the repeated Laplacian smoothing would make the features of connected graph nodes converge to similar vectors. This definition depicts only the degenerative behavior of nonlinear iteration in Fig. 6 (c) and (d). Obviously, the entanglement of layer weights and message propagations in nonlinear GNNs (e.g., SAGE and GCN) ruins the formation of eigen-subspaces. Over-smoothing, in this case, reflects the information decay from the starting iteration space to the fixed point. In Fig. 6 (c), SAGE (nonlinear DA) converges to the dominant eigenvector $1_n$; in Fig. 6 (d), GCN converges to the dominant eigenvector of $I + D^{-\frac{1}{2}}AD^{-\frac{1}{2}}$, that is, the square root of the node degree—$D^{-\frac{1}{2}}1_n$. The linear GCN (DAD) performs similarly to DA and outperforms GCN significantly in deep iterations. Table 3 lists the node and feature indiscernibility at the 50th iteration for SAGE and GCN. The statistics of the 1000th iteration for DA, RWN, RAN, and DAD are listed in Table 4. The nodes and features of GCN and DAD have the lowest indiscernibility, consistent with the above fixed point analysis. ($D^{-\frac{1}{2}}1_n$ v.s. $1_n$).
 
\renewcommand{\thetable}{\arabic{table}}
\begin{table}
\caption{Statistics of nonlinear models at the 50th iteration.}
\label{table-3}
\centering

\setlength{\tabcolsep}{6.52mm}{
\begin{tabular}{ccc}
\hline
\cellcolor[HTML]{FFFFFF}{\color[HTML]{333333} \textbf{Indiscernibility}} & \textbf{SAGE} & \textbf{GCN} \\ \hline
Node Level                                                               & 100\%         & 52\%         \\
Feature Level                                                            & 100\%         & 45\%         \\ \hline
\end{tabular}
}
\end{table}

\renewcommand{\thetable}{\arabic{table}}
\begin{table}
\caption{Statistics of linear models at the 1000th iteration.}
\label{table-4}
\centering

\setlength{\tabcolsep}{4.22mm}{
\begin{tabular}{cllll}
\hline
\multicolumn{1}{l}{\cellcolor[HTML]{FFFFFF}{\color[HTML]{333333} \textbf{Indiscernibility}}} & \textbf{DA}   & \textbf{RWN} & \textbf{RAN}  & \textbf{DAD}  \\ \hline
Node Level                                                                                   & 91\%          & 81\%         & 90\%          & 77\%          \\
Feature Level                                                                                & 25\%          & 92\%         & 26\%          & 33\%          \\
\textbf{Maximum}                                                                             & \textbf{91\%} & \textbf{92}  & \textbf{90\%} & \textbf{77\%} \\ \hline
\end{tabular}
}
\end{table}

The explorations in linear, nonlinear, and irregular iterations confirm our following conclusions:
  
  \begin{itemize}
	\item [1.]
	 The formation, duration, and collapse of the eigen-subspace are three stages of message iteration.
	 \item [2.] 
	 Over-smoothing is the information decay among the starting iteration space, eigen-subspace, and fixed point.
	 \end{itemize}

\subsection{Analysis and Exploration of Attractive Models}

\vspace{0.3cm}
\noindent \textit{4.3.1 Analysis and Verification of Linear Stress Models}
\vspace{0.1cm}

Specifying only local distances for node pairs is not sufficient for avoiding the problem often observed in the high-dimensional embedding algorithm and the classical graph drawing, where multiple nodes share the same position \cite{ref36}. The same is true in graph neural networks. However, over-smoothing does not conflict with the construction of deep models. As shown in Fig. 7, local distance-based StressDA and S-StressDA can alleviate and even prevent the performance deterioration of DA. The performance of StressDA fluctuates obviously after the 75th iteration, while S-StressDA shows superior performance in all iterations. Even after 10000 iterations, the results of S-StressDA will still be so excellent. The reason lies in its weight. $X^{(k)} = (1-\alpha)MX^{(k-1)} + \alpha X_0$ can be rewritten as 
\begin{equation}
\begin{aligned}
X^{(k)} = &\alpha X_0 + \alpha(1-\alpha)MX_0 +  \alpha(1-\alpha)^2M^2X_0+...\\ &+\alpha(1-\alpha)^kM^kX_0	
\end{aligned}
\end{equation}

S-StressDA essentially performs a linear combination of the intermediate results. The sum of the coefficients is $1-(1-\alpha)^k$. With increasing $k$, the importance of the new iteration output decreases dramatically. When setting $\alpha=0.1$ and $k=1,2,...,100$, as shown in Fig. 8, about 99.2\% of the output information has been determined at the 45th iteration. The coefficient of a single iteration, $\alpha(1-\alpha)^{k-1}$, exponentially decreases with the increase of $k$. \textbf{Therefore, building very deep models may be meaningless since the model information will no longer change with the iteration depth.}

\begin{figure}[!t]
\centering
\includegraphics[width=3.5in, height = 1.5in]{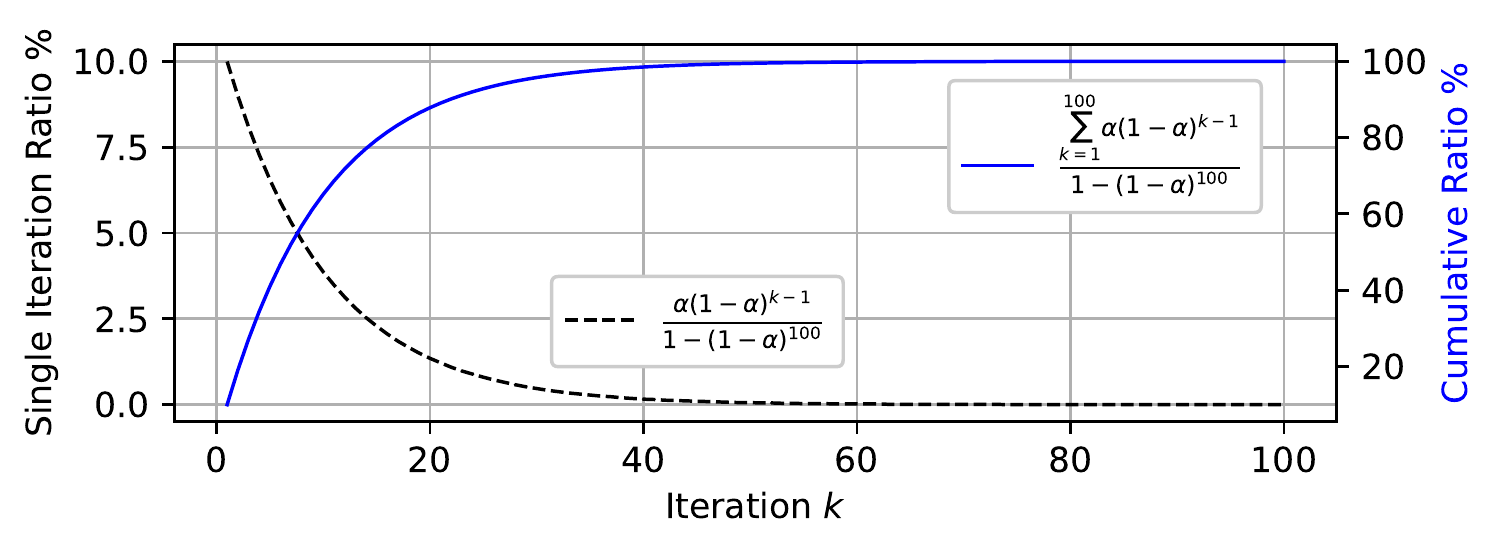}
% where an -eps-converted-to.pdf filename suffix will be assumed under latex, 
% and a .pdf suffix will be assumed for pdflatex; or what has been declared
% via \DeclareGraphicsExtensions.
\caption{Statistics of the iteration importance.}
%\label{fig_sim}
\end{figure}

 \renewcommand{\thetable}{\arabic{table}}
\begin{table}
\caption{Summary of classification accuracy (\%) results on citation networks.}
\label{table-5}
\centering

\setlength{\tabcolsep}{2.7mm}{
\begin{tabular}{llll}
\hline
\textbf{Models} & \textbf{Cora}            & \textbf{CiteSeer}      & \textbf{PubMed}         \\ \hline
DA              & 82.2±0.2\%               & 72.1±0.4\%             & 80.5±0.6\%              \\
StressDA        & 82.3±0.4\%               & 71.9±0.6\%             & 79.6±0.3\%              \\
S-StressDA      & 83.1±0.5\%               & 72.5±0.3\%             & 81.7±0.3\%              \\
DAD             & 82.5±0.4\%               & 72.4±0.2\%             & 81.9±0.5\%              \\
StressDAD       & 82.5±0.7\%               & 72.6±0.5\%             & 81.4±0.3\%              \\
S-StressDAD     & \textbf{84.2±0.3\%(65)} & \textbf{72.7±0.2\%(2)} & \textbf{82.2±0.1\%(15)} \\ \hline
\end{tabular}
}
\end{table}

% \usepackage{multirow}
\renewcommand{\thetable}{\arabic{table}}

\begin{table}
\caption{Node classification accuracy (\%) results for the model depth test.}
\label{table-6}
\centering

\setlength{\tabcolsep}{0.95mm}{
\begin{tabular}{ll|llllll}
\hline
                                   &                                          & \multicolumn{6}{c}{\textbf{Layers}}                                                                                                                                                                                                                             \\
\multirow{-2}{*}{\textbf{Dataset}} & \multirow{-2}{*}{\textbf{Model}}         & \multicolumn{1}{c}{\textbf{2}}                             & \multicolumn{1}{c}{\textbf{4}} & \multicolumn{1}{c}{\textbf{8}}         & \multicolumn{1}{c}{\textbf{16}}        & \multicolumn{1}{c}{\textbf{32}}        & \multicolumn{1}{c}{\textbf{64}}        \\ \hline
                                   & GCN                                      & \multicolumn{1}{c}{\textbf{81.93}}                         & 80.01                          & 77.02                                  & 22.14                                  & 23.34                                  & 22.11                                  \\
                                   & \cellcolor[HTML]{FFFA70}ResGCN           & \multicolumn{1}{c}{\cellcolor[HTML]{FFFA70}-}              & \cellcolor[HTML]{FFFA70}80.03  & \cellcolor[HTML]{FFFA70}68.91          & \cellcolor[HTML]{FFFA70}21.84          & \cellcolor[HTML]{FFFA70}21.73          & \cellcolor[HTML]{FFFA70}21.85          \\
                                   & \cellcolor[HTML]{FFFA70}Initial-ResGCN   & \multicolumn{1}{c}{\cellcolor[HTML]{FFFA70}-}              & \cellcolor[HTML]{FFFA70}80.72  & \cellcolor[HTML]{FFFA70}80.12          & \cellcolor[HTML]{FFFA70}80.30          & \cellcolor[HTML]{FFFA70}81.13          & \cellcolor[HTML]{FFFA70}80.72          \\
                                   & \cellcolor[HTML]{FFFA70}GCN(group\_norm) & \cellcolor[HTML]{FFFA70}\textbf{82.22}                     & \cellcolor[HTML]{FFFA70}79.84  & \cellcolor[HTML]{FFFA70}78.02          & \cellcolor[HTML]{FFFA70}67.14          & \cellcolor[HTML]{FFFA70}41.65          & \cellcolor[HTML]{FFFA70}24.94          \\
                                   & \cellcolor[HTML]{FFFA70}GCN(drop\_edge)  & \cellcolor[HTML]{FFFA70}\textbf{82.81}                     & \cellcolor[HTML]{FFFA70}82.04  & \cellcolor[HTML]{FFFA70}75.81          & \cellcolor[HTML]{FFFA70}75.75          & \cellcolor[HTML]{FFFA70}62.54          & \cellcolor[HTML]{FFFA70}49.56          \\
                                   & \cellcolor[HTML]{FFFA70}JKNetGCN         & \multicolumn{1}{c}{\cellcolor[HTML]{FFFA70}-}              & \cellcolor[HTML]{FFFA70}80.03  & \cellcolor[HTML]{FFFA70}80.52          & \cellcolor[HTML]{FFFA70}80.61          & \cellcolor[HTML]{FFFA70}81.23          & \cellcolor[HTML]{FFFA70}81.33          \\
                                   & \cellcolor[HTML]{22D2CA}DAGNN            & \cellcolor[HTML]{22D2CA}80.24                              & \cellcolor[HTML]{22D2CA}83.64  & \cellcolor[HTML]{22D2CA}\textbf{84.93} & \cellcolor[HTML]{22D2CA}84.41          & \cellcolor[HTML]{22D2CA}83.35          & \cellcolor[HTML]{22D2CA}83.06          \\
                                   & \cellcolor[HTML]{22D2CA}GCNII            & \cellcolor[HTML]{22D2CA}81.05                              & \cellcolor[HTML]{22D2CA}82.13  & \cellcolor[HTML]{22D2CA}83.73          & \cellcolor[HTML]{22D2CA}84.44          & \cellcolor[HTML]{22D2CA}84.54          & \cellcolor[HTML]{22D2CA}\textbf{85.03} \\
\multirow{-9}{*}{Cora}             & \cellcolor[HTML]{FFCCC9}StressGCN        & \cellcolor[HTML]{FFCCC9}82.92                              & \cellcolor[HTML]{FFCCC9}84.62  & \cellcolor[HTML]{FFCCC9}84.62          & \cellcolor[HTML]{FFCCC9}\textbf{85.13} & \cellcolor[HTML]{FFCCC9}84.83          & \cellcolor[HTML]{FFCCC9}84.95          \\ \hline
                                   & GCN                                      & \textbf{72.12}                                             & 67.52                          & 62.02                                  & 34.43                                  & 20.81                                  & 19.85                                  \\
                                   & \cellcolor[HTML]{FFFA70}ResGCN           & \multicolumn{1}{c}{\cellcolor[HTML]{FFFA70}-}              & \cellcolor[HTML]{FFFA70}66.83  & \cellcolor[HTML]{FFFA70}63.63          & \cellcolor[HTML]{FFFA70}21.25          & \cellcolor[HTML]{FFFA70}21.08          & \cellcolor[HTML]{FFFA70}21.23          \\
                                   & \cellcolor[HTML]{FFFA70}Initial-ResGCN   & \multicolumn{1}{c}{\cellcolor[HTML]{FFFA70}-}              & \cellcolor[HTML]{FFFA70}69.02  & \cellcolor[HTML]{FFFA70}67.61          & \cellcolor[HTML]{FFFA70}68.93          & \cellcolor[HTML]{FFFA70}69.15          & \cellcolor[HTML]{FFFA70}68.26          \\
                                   & \cellcolor[HTML]{FFFA70}GCN(group\_norm) & \multicolumn{1}{c}{\cellcolor[HTML]{FFFA70}\textbf{72.43}} & \cellcolor[HTML]{FFFA70}66.81  & \cellcolor[HTML]{FFFA70}63.75          & \cellcolor[HTML]{FFFA70}48.04          & \cellcolor[HTML]{FFFA70}35.83          & \cellcolor[HTML]{FFFA70}19.23          \\
                                   & \cellcolor[HTML]{FFFA70}GCN(drop\_edge)  & \cellcolor[HTML]{FFFA70}\textbf{72.31}                     & \cellcolor[HTML]{FFFA70}70.62  & \cellcolor[HTML]{FFFA70}61.42          & \cellcolor[HTML]{FFFA70}57.21          & \cellcolor[HTML]{FFFA70}41.65          & \cellcolor[HTML]{FFFA70}34.42          \\
                                   & \cellcolor[HTML]{FFFA70}JKNetGCN         & \multicolumn{1}{c}{\cellcolor[HTML]{FFFA70}-}              & \cellcolor[HTML]{FFFA70}71.31  & \cellcolor[HTML]{FFFA70}70.03          & \cellcolor[HTML]{FFFA70}70.34          & \cellcolor[HTML]{FFFA70}71.02          & \cellcolor[HTML]{FFFA70}71.57          \\
                                   & \cellcolor[HTML]{22D2CA}DAGNN            & \cellcolor[HTML]{22D2CA}71.71                              & \cellcolor[HTML]{22D2CA}72.53  & \cellcolor[HTML]{22D2CA}72.91          & \cellcolor[HTML]{22D2CA}73.23          & \cellcolor[HTML]{22D2CA}\textbf{73.35} & \cellcolor[HTML]{22D2CA}72.18          \\
                                   & \cellcolor[HTML]{22D2CA}GCNII            & \cellcolor[HTML]{22D2CA}68.20                              & \cellcolor[HTML]{22D2CA}69.00  & \cellcolor[HTML]{22D2CA}70.33          & \cellcolor[HTML]{22D2CA}72.64          & \cellcolor[HTML]{22D2CA}73.27          & \cellcolor[HTML]{22D2CA}\textbf{73.69} \\
\multirow{-9}{*}{CiteSeer}         & \cellcolor[HTML]{FFCCC9}StressGCN        & \cellcolor[HTML]{FFCCC9}71.91                              & \cellcolor[HTML]{FFCCC9}73.30  & \cellcolor[HTML]{FFCCC9}\textbf{73.61} & \cellcolor[HTML]{FFCCC9}73.00          & \cellcolor[HTML]{FFCCC9}72.92          & \cellcolor[HTML]{FFCCC9}73.34          \\ \hline
                                   & GCN                                      & \textbf{81.01}                                             & 77.40                          & 72.84                                  & 38.01                                  & 39.60                                  & 38.82                                  \\
                                   & \cellcolor[HTML]{FFFA70}ResGCN           & \multicolumn{1}{c}{\cellcolor[HTML]{FFFA70}-}              & \cellcolor[HTML]{FFFA70}78.63  & \cellcolor[HTML]{FFFA70}67.92          & \cellcolor[HTML]{FFFA70}38.44          & \cellcolor[HTML]{FFFA70}38.81          & \cellcolor[HTML]{FFFA70}38.21          \\
                                   & \cellcolor[HTML]{FFFA70}Initial-ResGCN   & \multicolumn{1}{c}{\cellcolor[HTML]{FFFA70}-}              & \cellcolor[HTML]{FFFA70}78.27  & \cellcolor[HTML]{FFFA70}76.95          & \cellcolor[HTML]{FFFA70}77.04          & \cellcolor[HTML]{FFFA70}78.50          & \cellcolor[HTML]{FFFA70}74.32          \\
                                   & \cellcolor[HTML]{FFFA70}GCN(group\_norm) & \multicolumn{1}{c}{\cellcolor[HTML]{FFFA70}\textbf{81.90}} & \cellcolor[HTML]{FFFA70}78.14  & \cellcolor[HTML]{FFFA70}77.12          & \cellcolor[HTML]{FFFA70}73.83          & \cellcolor[HTML]{FFFA70}66.82          & \cellcolor[HTML]{FFFA70}63.03          \\
                                   & \cellcolor[HTML]{FFFA70}GCN(drop\_edge)  & \cellcolor[HTML]{FFFA70}79.62                              & \cellcolor[HTML]{FFFA70}79.44  & \cellcolor[HTML]{FFFA70}78.13          & \cellcolor[HTML]{FFFA70}78.51          & \cellcolor[HTML]{FFFA70}77.00          & \cellcolor[HTML]{FFFA70}61.55          \\
                                   & \cellcolor[HTML]{FFFA70}JKNetGCN         & \multicolumn{1}{c}{\cellcolor[HTML]{FFFA70}-}              & \cellcolor[HTML]{FFFA70}80.63  & \cellcolor[HTML]{FFFA70}80.66          & \cellcolor[HTML]{FFFA70}\textbf{81.11} & \cellcolor[HTML]{FFFA70}80.50          & \cellcolor[HTML]{FFFA70}80.52          \\
                                   & \cellcolor[HTML]{22D2CA}DAGNN            & \cellcolor[HTML]{22D2CA}77.61                              & \cellcolor[HTML]{22D2CA}78.63  & \cellcolor[HTML]{22D2CA}80.42          & \cellcolor[HTML]{22D2CA}80.21          & \cellcolor[HTML]{22D2CA}\textbf{80.63} & \cellcolor[HTML]{22D2CA}80.23          \\
                                   & \cellcolor[HTML]{22D2CA}GCNII            & \cellcolor[HTML]{22D2CA}79.21                              & \cellcolor[HTML]{22D2CA}80.13  & \cellcolor[HTML]{22D2CA}79.84          & \cellcolor[HTML]{22D2CA}79.80          & \cellcolor[HTML]{22D2CA}\textbf{80.71} & \cellcolor[HTML]{22D2CA}80.44          \\
\multirow{-9}{*}{PubMed}           & \cellcolor[HTML]{FFCCC9}StressGCN        & \cellcolor[HTML]{FFCCC9}80.63                              & \cellcolor[HTML]{FFCCC9}81.60  & \cellcolor[HTML]{FFCCC9}81.54          & \cellcolor[HTML]{FFCCC9}\textbf{81.92} & \cellcolor[HTML]{FFCCC9}\textbf{81.92} & \cellcolor[HTML]{FFCCC9}81.81          \\ \hline
\end{tabular}
}
\end{table}

We also list the results of linear stress Laplacian models on citation benchmark networks in Table 5, where the best result is annotated with the model depth. Obviously, the more appropriate model depth choice depends on the target dataset.

\renewcommand{\thetable}{\arabic{table}}

\begin{table}
\caption{Comparison between StressGCN and previous deep models on ogbn-arxiv.}
\label{table-7}
\centering

\setlength{\tabcolsep}{7.02mm}{
\begin{tabular}{ccc}
\hline
{\color[HTML]{333333} \textbf{Model}} & {\color[HTML]{333333} \textbf{Val Acc(\%)}} & {\color[HTML]{333333} \textbf{Test Acc(\%)}} \\ \hline
{\color[HTML]{333333} DeeperGCN}      & {\color[HTML]{333333} 72.62±0.14\%}         & {\color[HTML]{333333} 71.92±0.16\%}          \\
{\color[HTML]{333333} DAGNN}          & {\color[HTML]{333333} 72.90±0.09\%}         & {\color[HTML]{333333} 72.09±0.25\%}          \\
{\color[HTML]{333333} GCNII}          & {\color[HTML]{333333} 73.50±0.08\%}         & {\color[HTML]{333333} 72.74±0.16\%}          \\
{\color[HTML]{333333} StressGCN}      & {\color[HTML]{333333} 74.00±0.06\%}         & {\color[HTML]{333333} 73.02±0.17\%}          \\ \hline
\end{tabular}
}
\end{table}

\vspace{0.3cm}
\noindent \textit{4.3.2 Analysis and Verification of Nonlinear Stress Models}
\vspace{0.1cm}

In Table 6, we can observe that on citation networks, the performance of GCN with Residual, GroupNorm, and DropEdge drops rapidly as the number of layers exceeds 8. GCN with InitialResidual and JKNet can prevent the performance deterioration of GCN but maintain a relatively low level of accuracy. StressGCN achieves the best results on most networks. Different from previous models, we emphasize the following two principles in building deep nonlinear models:

\begin{itemize}
	\item [1.]
	 Decoupling the entanglement of layer weights and message propagation.
	 \item [2.] 
	 Constructing an imbalanced combination of shallow iterations (in dominance) and deep iterations.
\end{itemize}

The most relevant models to our proposed principles are DAGNN and GCNII (PyTorch's built-in methods), which implement principles 1 and 2, respectively. As shown in Table 6, the performance of DAGNN and GCNII obviously improves as we increase the number of layers. Compared with StressGCN, GCNII always achieves its best result in the last few iterations. The combination of initial residual connection and identity mapping seems to hinder the emergence of good results in the early iterations of GCNII. As for DAGNN, its dense connection does not effectively balance shallow and deep iterations. StressGCN is the simplest and most effective structure we can extract from DAGNN and GCNII. Moreover, S-StressDAD achieves the best result on the PubMed dataset, outperforming nonlinear models by a noticeable margin.

In Table 7, we also compare StressGCN with DAGNN, GCNII, and DeeperGCN on ogbn-arxiv. All results confirmed the reliability and effectiveness of the two principles proposed in this paper.

\subsection{Analysis and Exploration of Repulsive Models}

\vspace{0.3cm}
\noindent \textit{4.4.1 Analysis and Verification of Sparse Repulsive GNNs}
\vspace{0.1cm}

Thus far, we can build arbitrarily deep GNNs under local message passing. However, without specifying distances for non-neighboring node pairs, above models will always tend to embed the nodes having similar neighbor structures into the same embedding vectors (unless one has a very good iteration initialization). We propose repulsive message passing to improve this defect of GNNs and verify its effectiveness in the following two sets of experiments.

\textbf{Link prediction}. Table 8 summarizes the performance of repulsive models (SR-GNNs and P-GNNs) and GCN on synthetic datasets with two kinds of graph features. When the graph feature is a 400-dimensional identity matrix, encoding node IDs and containing some position information, the performance difference between GCN and repulsive models on Grid-400 and Com-400 is not significant. When using the constant vector $1_n=[1,1,...,1]$ as the node feature, GCN can not distinguish structurally isomorphic nodes in Grid-1 and Com-1. The constant graph feature test is more challenging than the random graph feature test (presented in Section 4.2) for message passing in that even power iteration requires the initial iteration vector must be non-constant. Therefore, $1_n$ inevitably ruins the performance of GCN.

While SR-GNN and P-GNN can recognize node positions by their different distances to pivots (or anchor sets) and outperform GCN by a large margin. There is no clear difference between P-GNN and SR-GNN in the performance. Their fast mode (e.g., SR-GNN-F and P-GNN-F) sometimes can serve as a good substitute. By comparing the static and dynamic modes of SR-GNN-F, we can find that it is not economical to sample pivot nodes in each forward pass.

\textbf{Pairwise node classification}. Table 9 summarizes the performance of repulsive models and GCN on real datasets. Repulsive models outperform the standard GCN by a large margin. The conclusions about SR-GNN and P-GNN are almost consistent with the conclusions in link prediction tasks. Although we emphasize the importance of repulsion in recognizing the global node positions, it does not mean that repulsive information is more important than the neighborhood structure or attractive information. As shown in Table 9, except for the Email dataset, which is equipped with a constant feature $1_n$, the performance of neighborhood aggregation (GCNII and StressGCN) is better than that of repulsive models. The choice of the message-passing scheme thus depends on the problem to be solved and the available information in the graph feature. Moreover, although the repulsion in SR-GNN-R is more approximate to full stress, SR-GNN-E and SR-GNN-F perform better on most datasets and are much more economical.

\renewcommand{\thetable}{\arabic{table}}

\begin{table}
\caption{SR-GNNs compared to P-GNNs on link prediction tasks, measured in ROC AUC.}
\label{table-8}
\centering

\setlength{\tabcolsep}{1.2mm}{
\begin{tabular}{ccccc}
\hline
\textbf{Model}    & \textbf{Grid-400}                                         & \textbf{Com-400} & \textbf{Grid-1}                                           & \textbf{Com-1}  \\ \hline
GCN               & \multicolumn{1}{l}{\textbf{0.682±0.030}}                & \textbf{0.982±0.005}   & \multicolumn{1}{l}{\textbf{0.469±0.006}}                & \textbf{0.484± 0.019} \\
\rowcolor[HTML]{FFFA70} 
P-GNN-E           & \multicolumn{1}{l}{\cellcolor[HTML]{FFFA70}0.694±0.033} & 0.975±0.007            & \multicolumn{1}{l}{\cellcolor[HTML]{FFFA70}0.940±0.027} & 0.971± 0.008          \\
\rowcolor[HTML]{FFFA70} 
SRGNN-E           & \textbf{0.758±0.42}                                     & 0.978±0.004            & \textbf{0.947±0.027}                                    & 0.979±0.005           \\
\rowcolor[HTML]{22D2CA} 
P-GNN-F           & 0.600±0.037                                             & 0.963±0.009            & 0.685±0.078                                             & 0.978±0.005           \\
\rowcolor[HTML]{22D2CA} 
SR-GNN-F(S)  & 0.540±0.080                                             & 0.978±0.003            & 0.781±0.021                                             & 0.979±0.008           \\
\rowcolor[HTML]{22D2CA} 
SR-GNN-F(D) & 0.536±0.050                                             & 0.977±0.006            & 0.788±0.040                                             & 0.981±0.007           \\
SR-GNN-R          & 0.707±0.035                                             & \textbf{0.981±0.005}   & 0.696±0.040                                             & \textbf{0.983±0.004}  \\ \hline
\end{tabular}
}
\end{table}

\renewcommand{\thetable}{\arabic{table}}

\begin{table}
\caption{Performance on pairwise node classification tasks, measured in ROC AUC.}
\label{table-9}
\centering

\setlength{\tabcolsep}{1.51mm}{
\begin{tabular}{cllll}
\hline
\textbf{Model} & \textbf{Cornell}     & \textbf{Email}       & \textbf{Texas}       & \textbf{Wisconsin}   \\ \hline
GCN            & 0.800±0.020          & 0.535±0.028          & 0.815±0.017          & 0.829±0.016          \\
\rowcolor[HTML]{FFFA70} 
P-GNN-E        & 0.854±0.014          & 0.689±0.032          & 0.851±0.012          & 0.8512±0.012         \\
\rowcolor[HTML]{FFFA70} 
SR-GNN-E       & 0.841±0.023          & \textbf{0.802±0.020} & 0.847±0.019          & 0.824±0.017          \\
\rowcolor[HTML]{22D2CA} 
P-GNN-F        & \textbf{0.860±0.015} & 0.677±0.031          & 0.861±0.010          & \textbf{0.855±0.016} \\
\rowcolor[HTML]{22D2CA} 
SR-GNN-F       & 0.854±0.013          & 0.767±0.021          & \textbf{0.867±0.011} & 0.843±0.019          \\
SR-GNN-R       & 0.841±0.026          & 0.785±0.025          & 85.8±0.013           & 0.830±0.014          \\ \hline
                                                           
GCNII          & 0.871±0.016          & 0.694±0.006          & 0.866±0.014          & 0.890±0.009          \\
StressGCN      & \textbf{0.873±0.013} & \textbf{0.694±0.006} & \textbf{0.871±0.008} & \textbf{0.895±0.005} \\ \hline
\end{tabular}
}
\end{table}

\begin{figure}[h]

\begin{minipage}{0.48\linewidth}
 	\vspace{3pt}
 	\centerline{\includegraphics[width=1.3in]{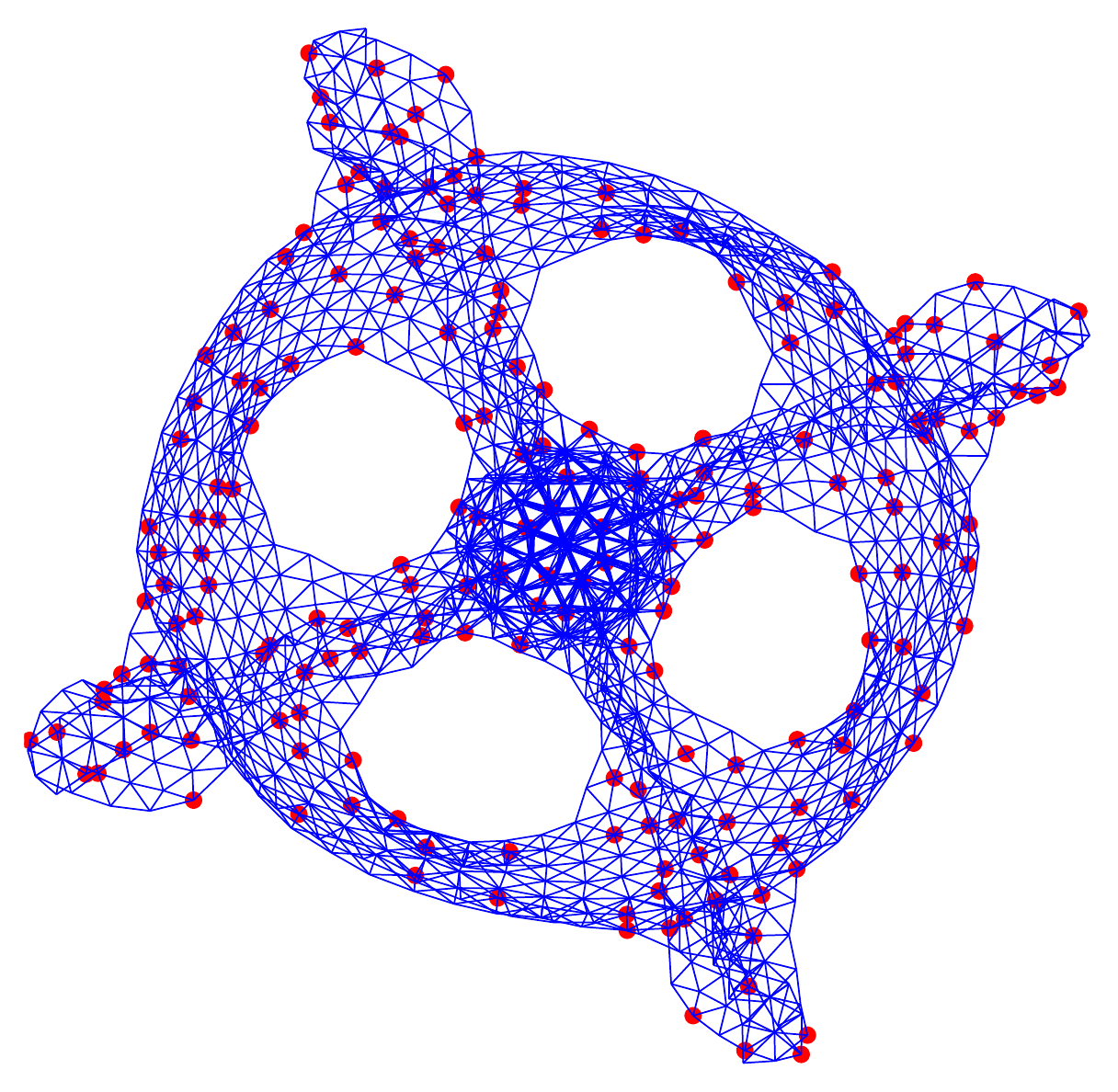}}
 	%\vspace{3pt}
 	\centerline{(a)}
 	\centerline{\includegraphics[width=1.4in, angle=168]{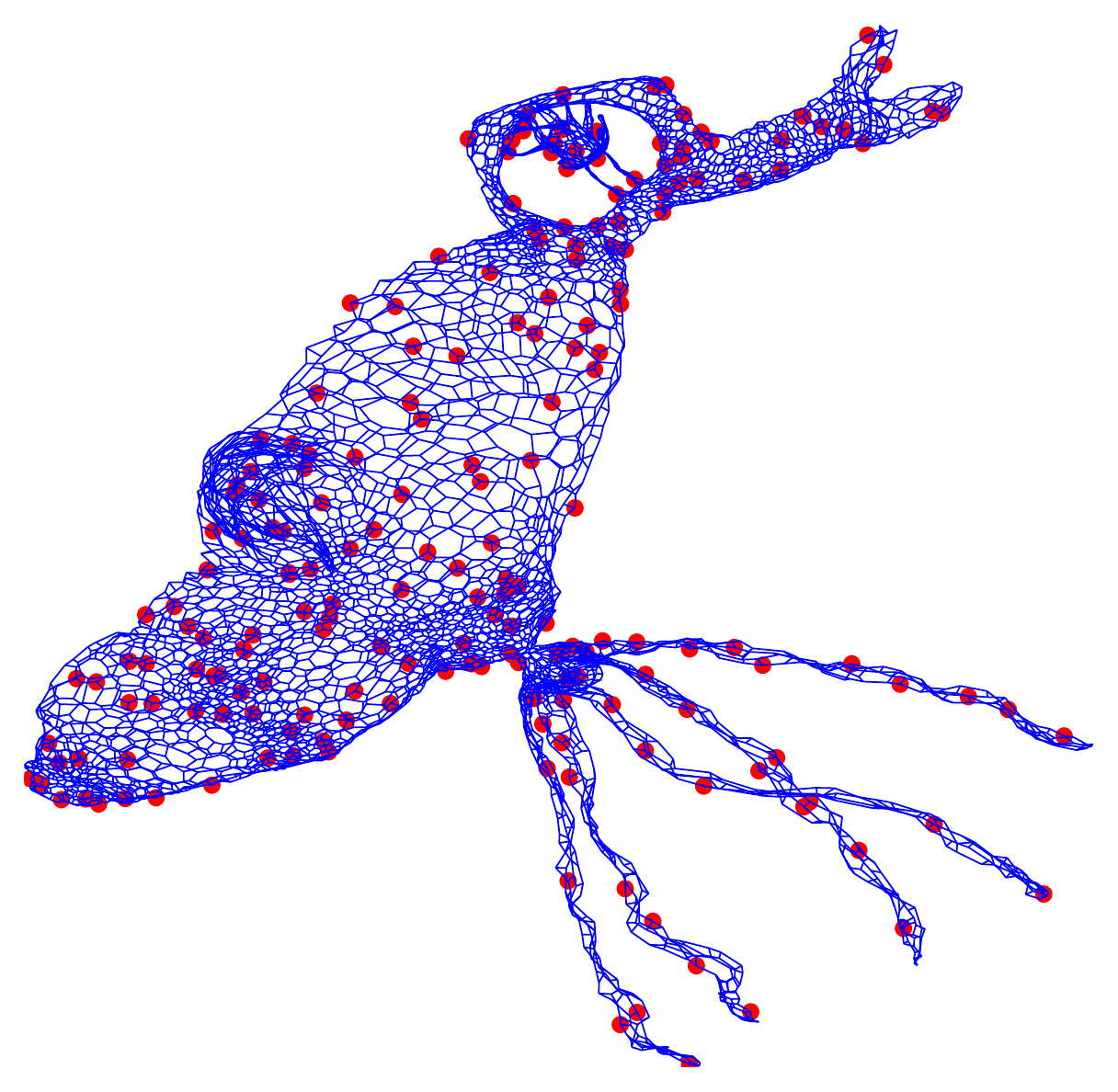}}
 	%\vspace{3pt}
 	\centerline{(c)}
 \end{minipage}
 \begin{minipage}{0.48\linewidth}
 	\vspace{3pt}
 	\centerline{\includegraphics[width=1.3in]{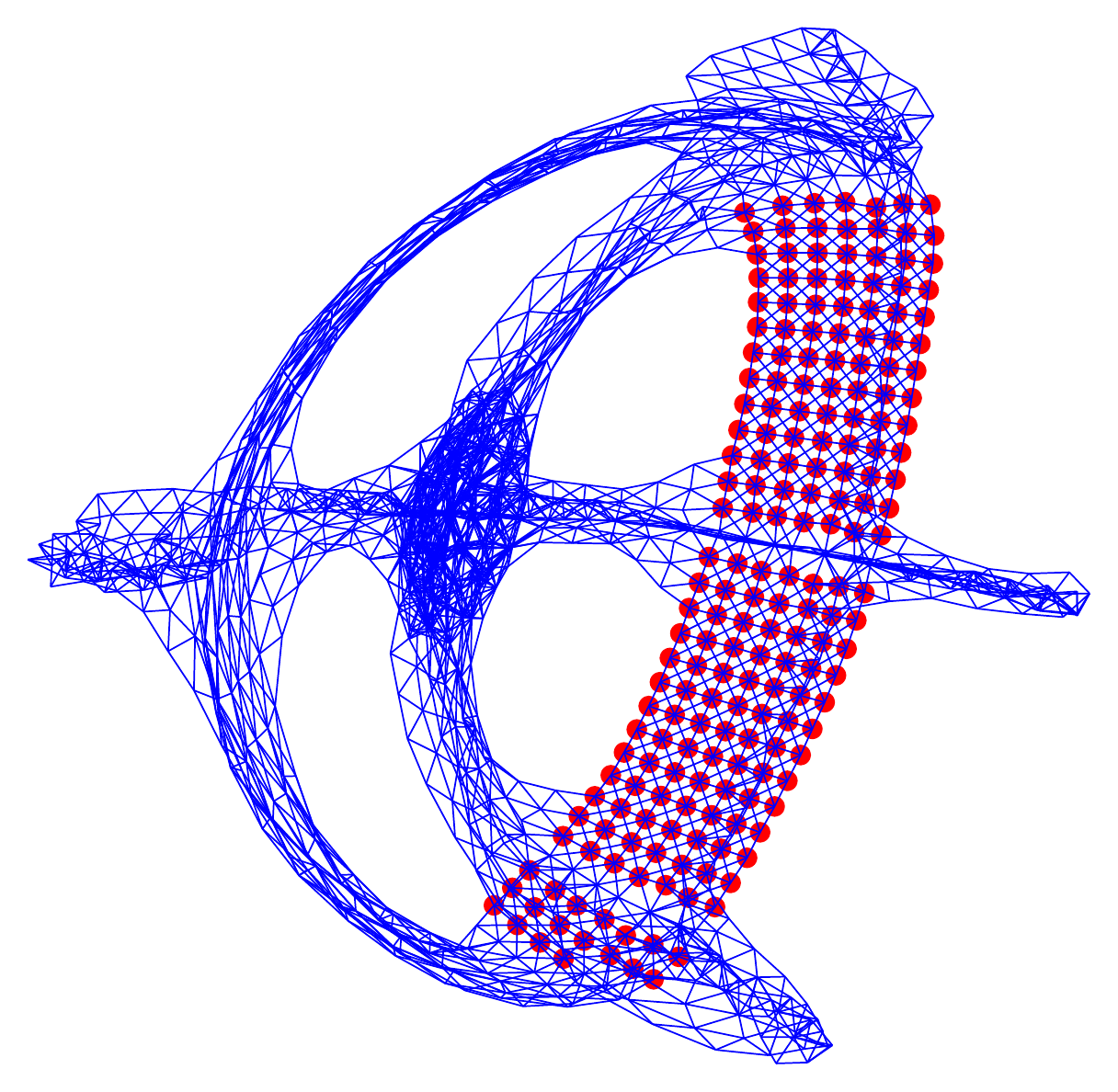}}
 	%\vspace{3pt}
 	\centerline{(b)}
 	\centerline{\includegraphics[width=1.4in, angle=168]{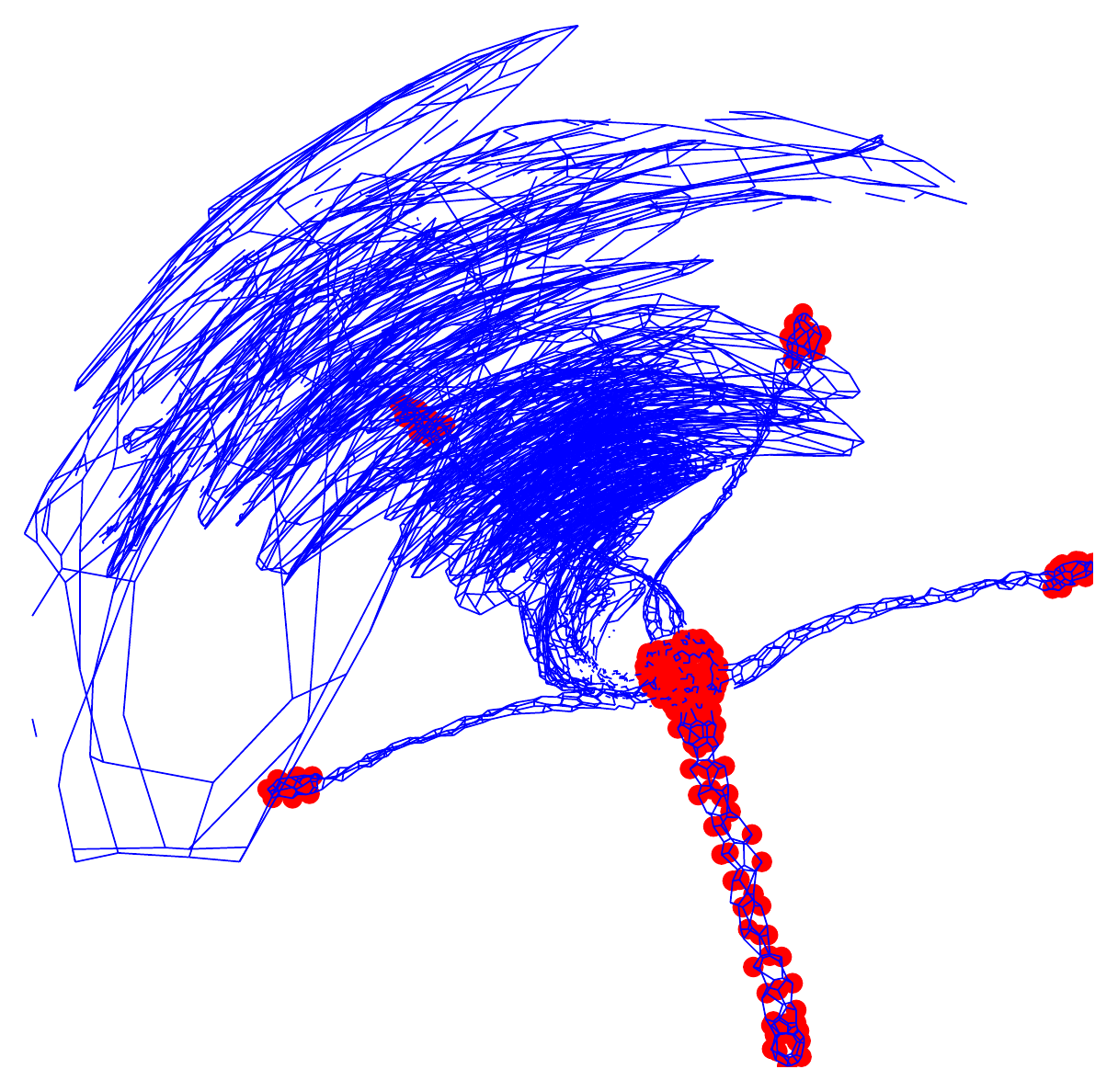}}
 	%\vspace{3pt}
 	\centerline{(d)}
\end{minipage}

\caption{Graph visualization of different pivot selection strategies. From top to bottom: dwt\_1005 (1,005 vertices) and commanche\_dual (7,920). Each graph is approximated with 200 pivots, and red nodes represent pivots.}
\label{Fig.9}
\end{figure}

We are also concerned with evaluating the impact of different pivot sampling strategies in graph visualization. HDE-Pivot and P-GNN-Anchor selection strategies are used to choose 200 pivot nodes for the sparse stress drawing, respectively. In Fig.9 (a) and (c), we can see that the pivots generated via HDE-Pivot are well distributed over the graph, creating regions of equal complexity, and are central in the drawing of their regions. While for the pivots generated by P-GNN-Anchor, the pivot nodes are not evenly distributed in the graph, resulting in varying degrees of graph layout distortion. When the graph structure is relatively symmetric and not very complex (e.g., dwt\_1005), we can still recognize the original graph structure. For the graph having an abstract structure like the helicopter in Fig. 9 (c), the ill-selected pivots of P-GNN-Anchor, residing in the blades, severely ruin the sparse stress layout (Fig. 9 (d)).

\begin{figure*}[h]
	\begin{minipage}{0.16\linewidth}
		\vspace{3pt}
        %这个图片路径替换成你的图片路径即可使用
		\centerline{\includegraphics[width=1.1in]{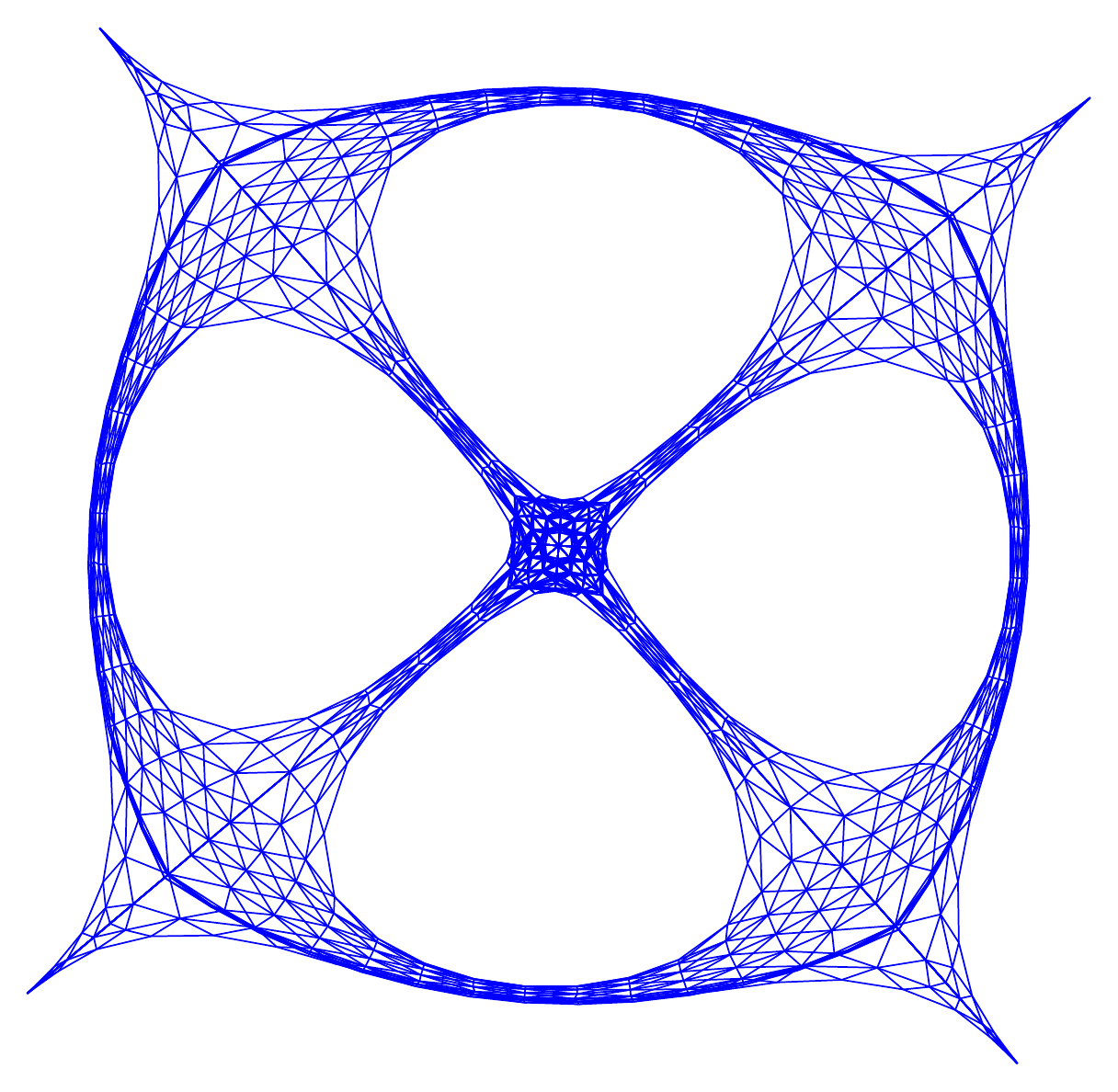}}
		\vspace{3pt}
 		\centerline{\includegraphics[width=1.1in, angle=10]{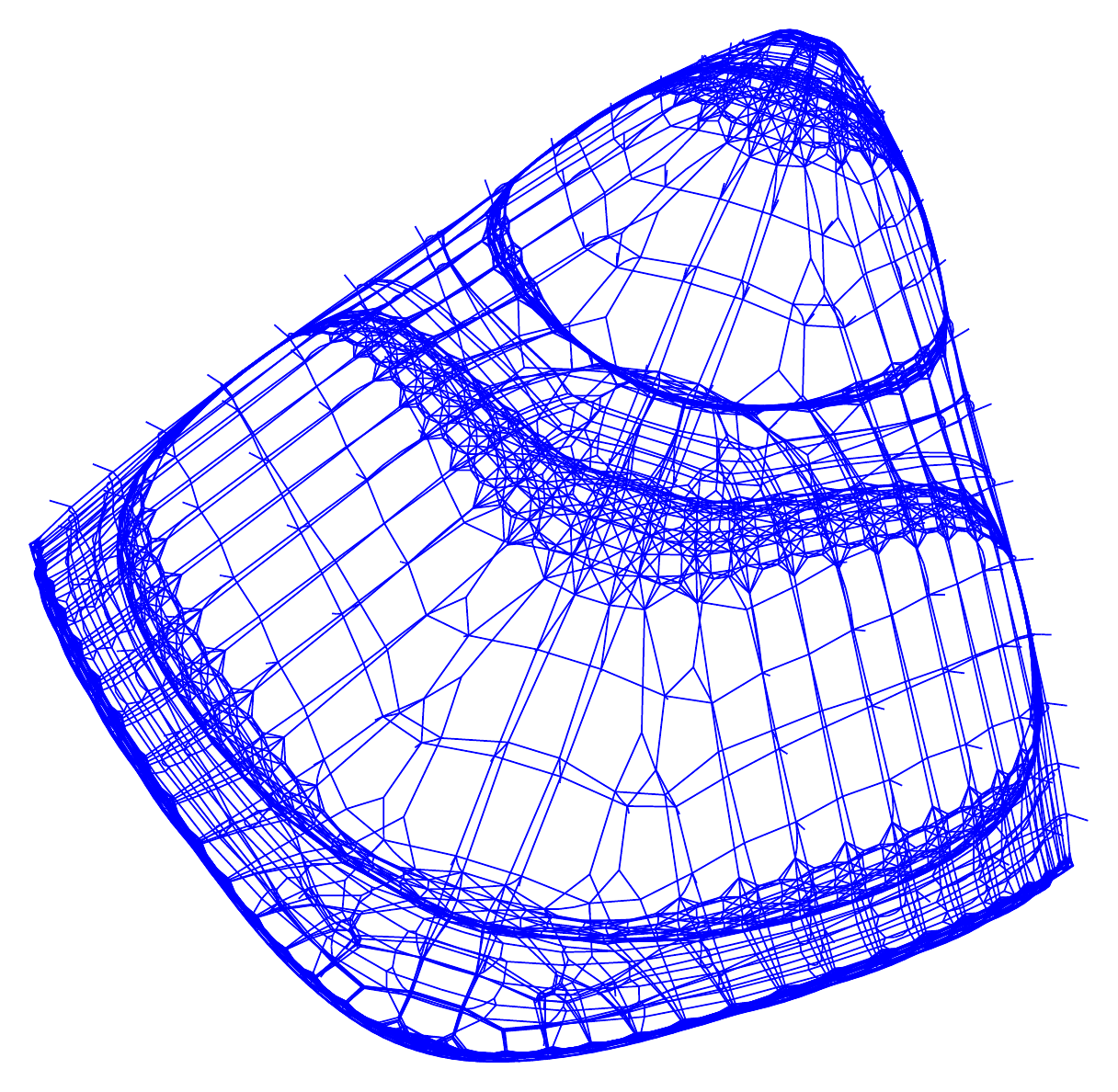}}
          % 加入对这列的图片说明
		\centerline{vp-0}
	\end{minipage}
	\begin{minipage}{0.16\linewidth}
		\vspace{3pt}
		\centerline{\includegraphics[width=1.2in]{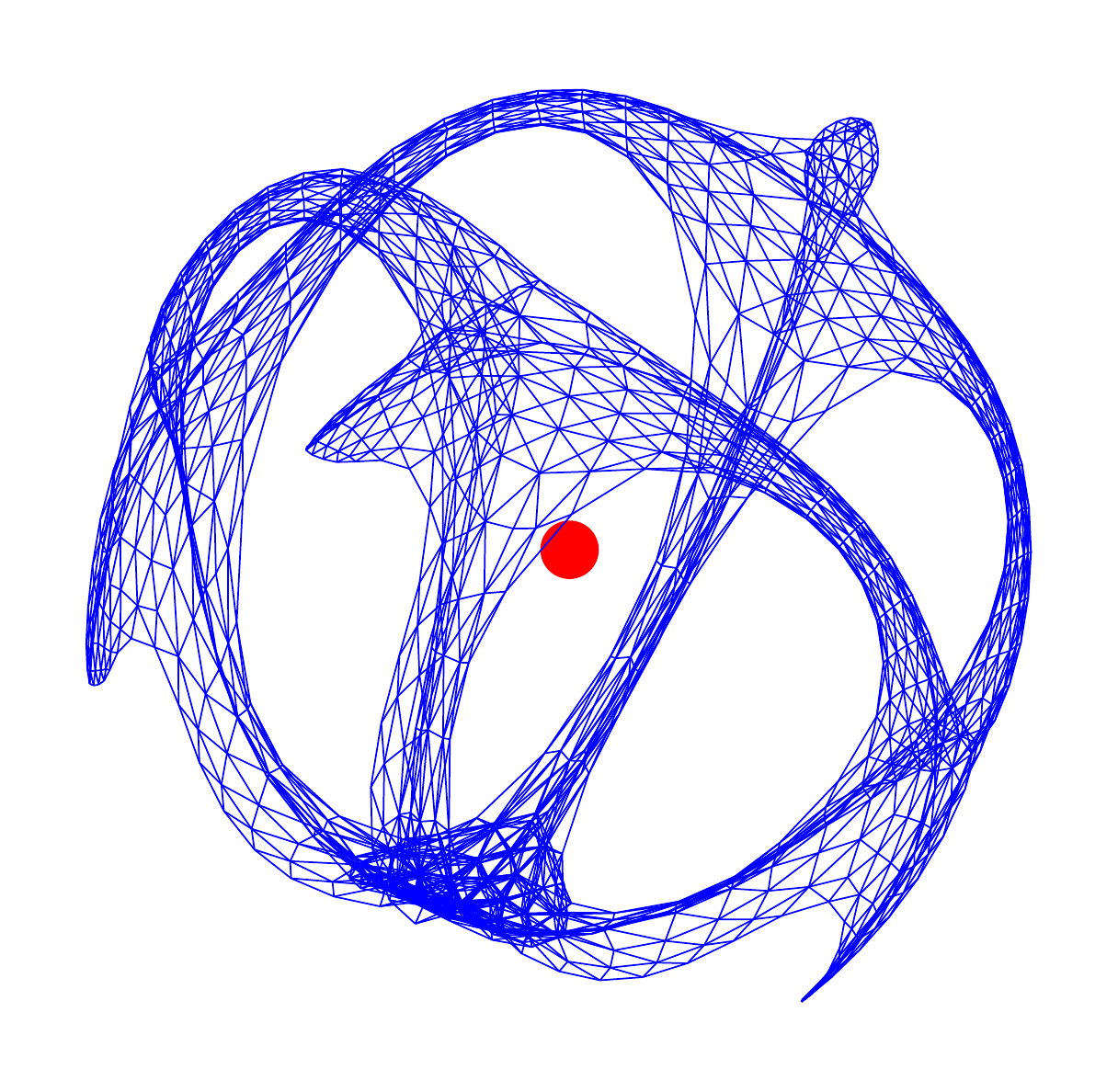}}
		\vspace{3pt}
 		\centerline{\includegraphics[width=1.2in]{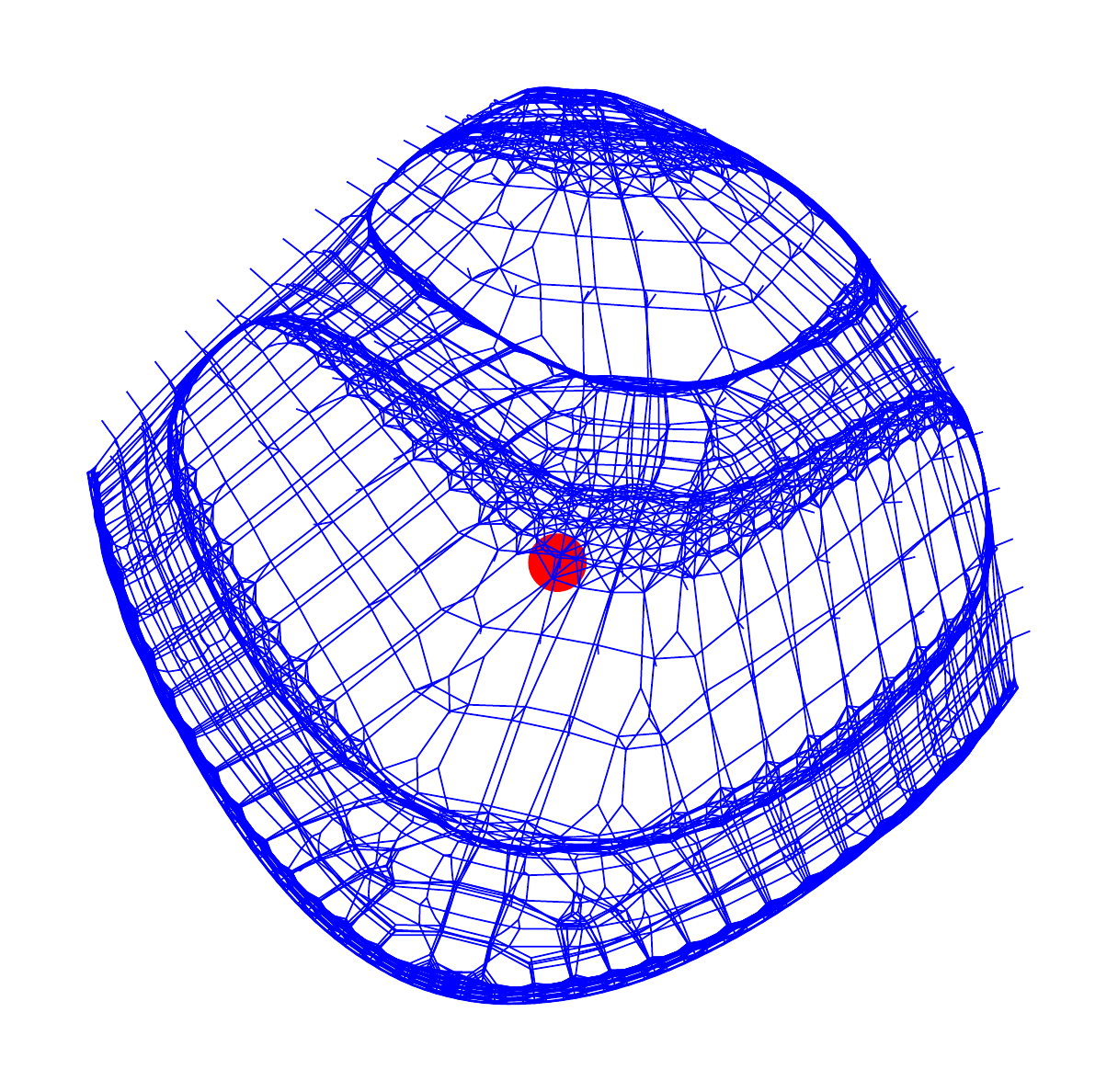}}
		\centerline{vp-1}
	\end{minipage}
	\begin{minipage}{0.16\linewidth}
		\vspace{3pt}
		\centerline{\includegraphics[width=1.2in]{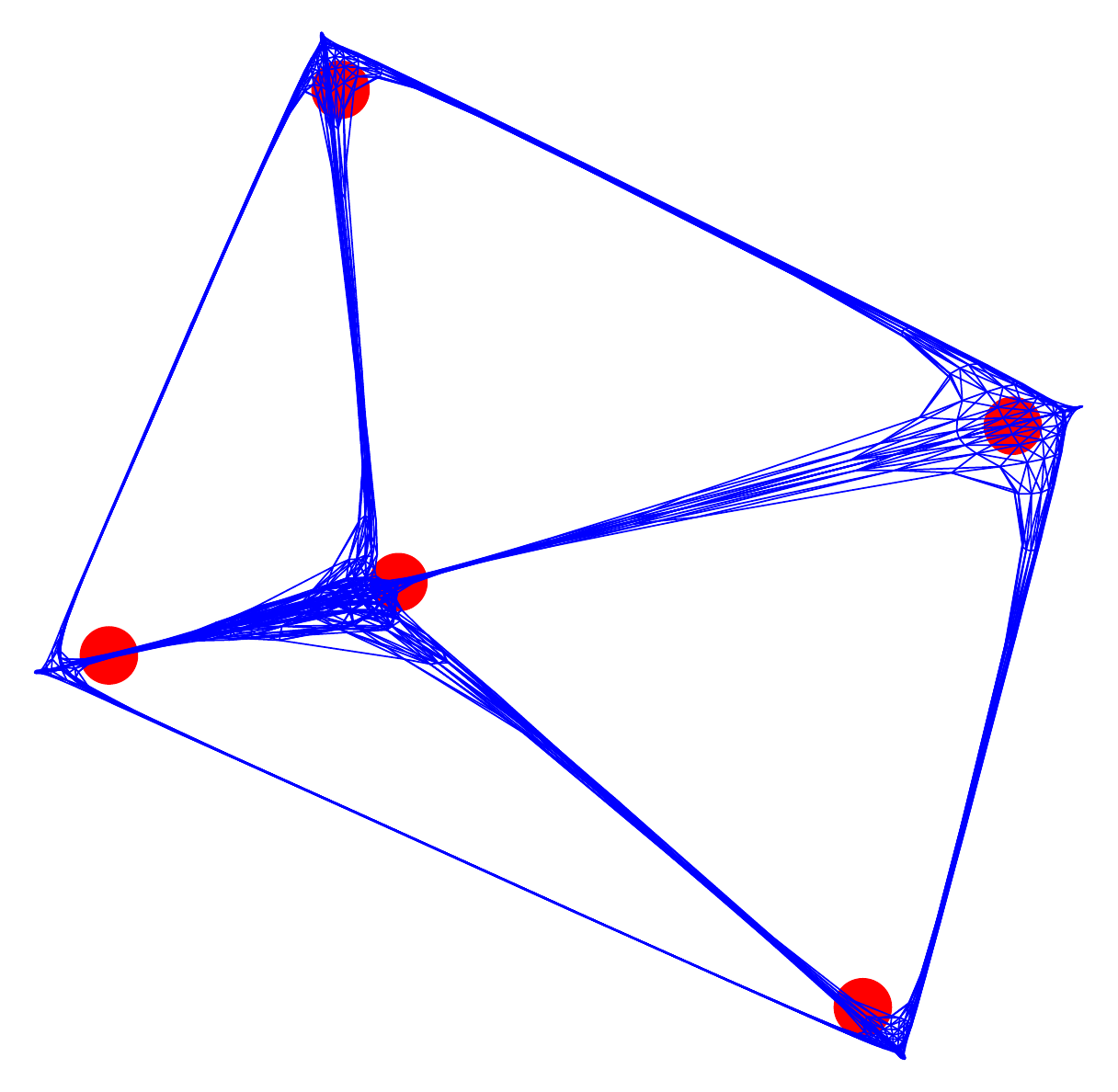}}
		\vspace{3pt}
 		\centerline{\includegraphics[width=1.2in]{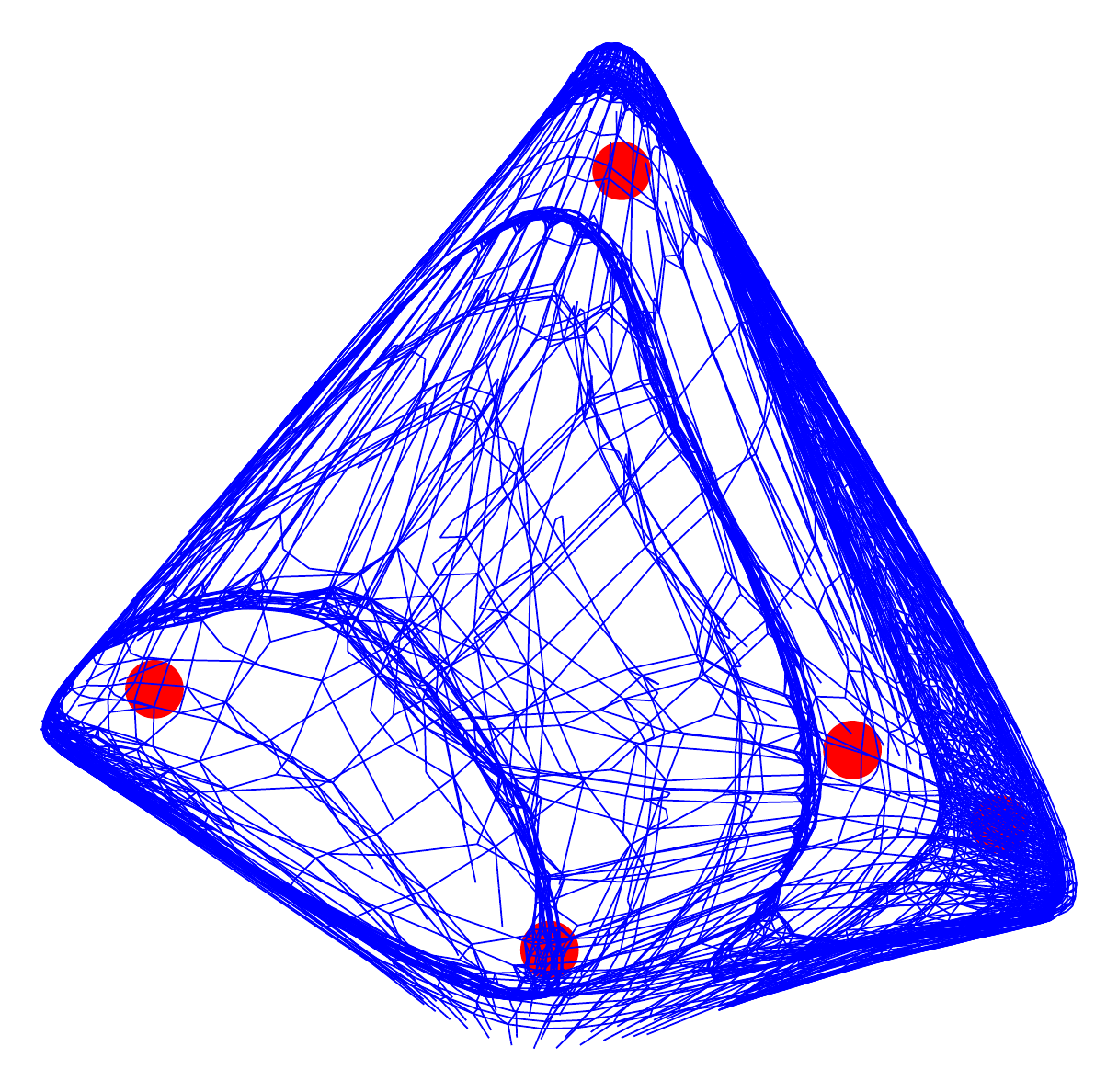}}
		\centerline{vp-5}
	\end{minipage}
	\begin{minipage}{0.16\linewidth}
		\vspace{3pt}
		\centerline{\includegraphics[width=1.1in, angle=-10]{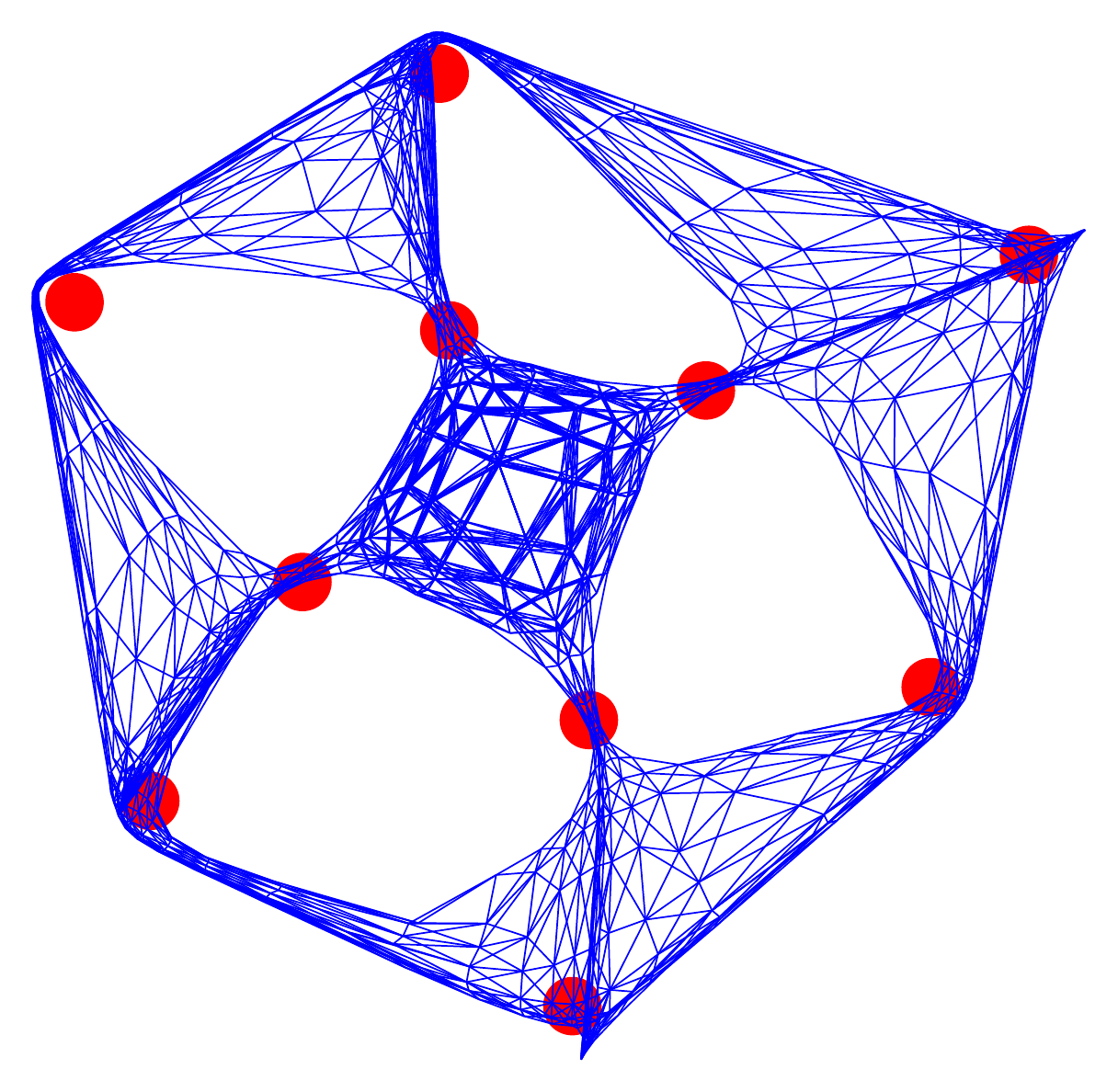}}
		\vspace{3pt}
 		\centerline{\includegraphics[width=1.1in]{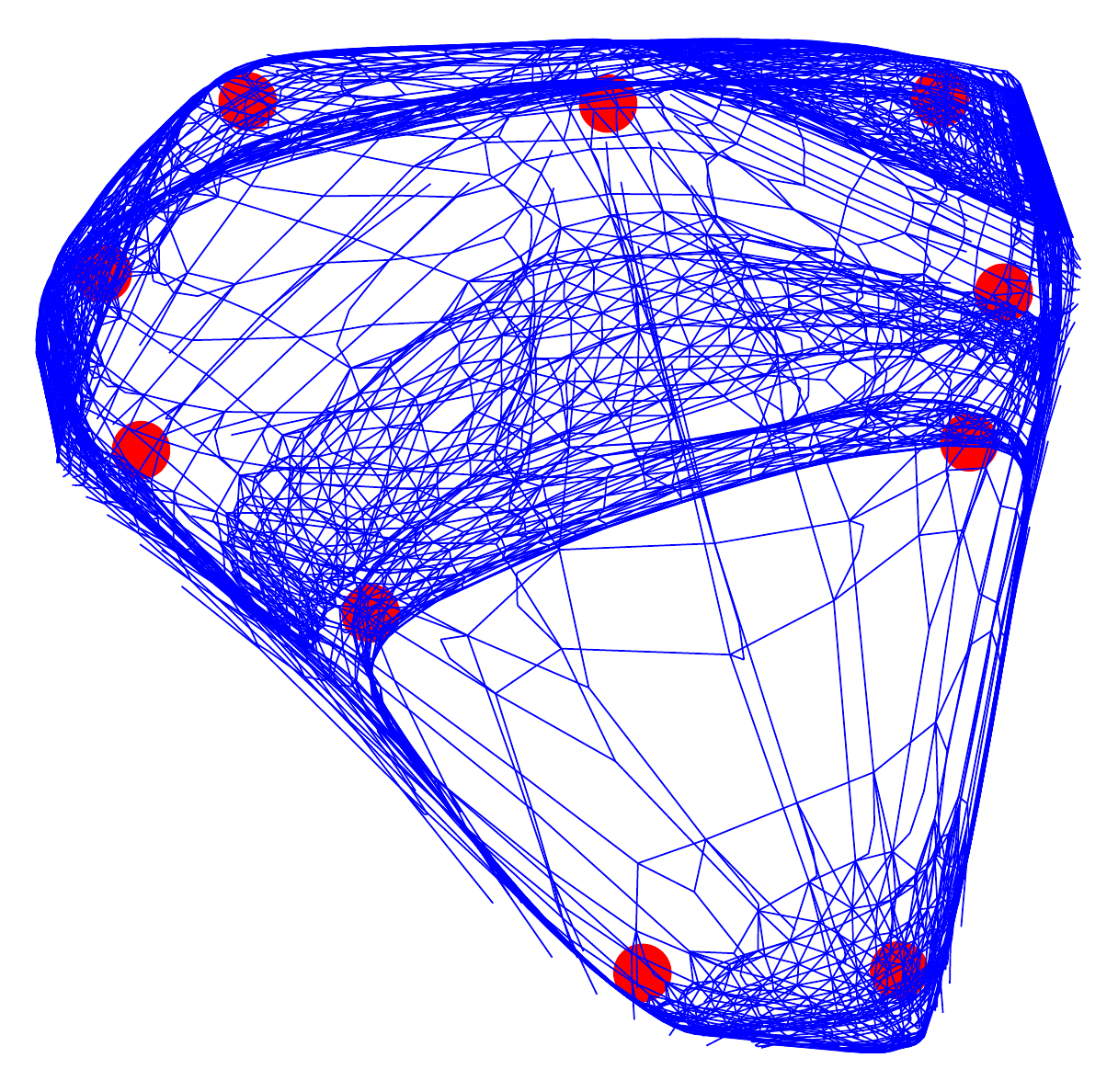}}
 	
		\centerline{vp-10}
	\end{minipage}	
	\begin{minipage}{0.16\linewidth}
		\vspace{3pt}
		\centerline{\includegraphics[width=1.2in]{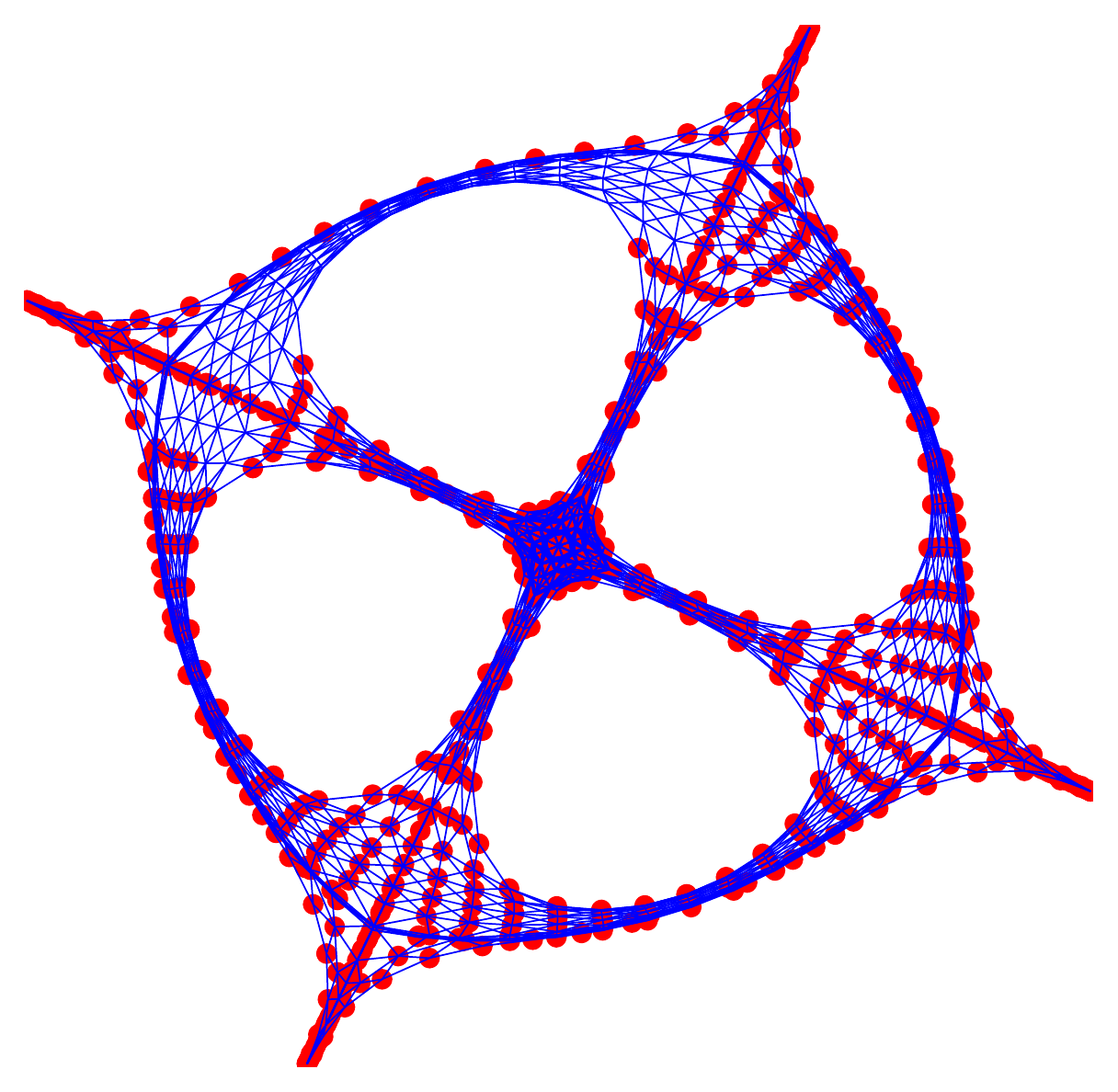}}
		\vspace{3pt}
 		\centerline{\includegraphics[width=1.1in, angle=-9]{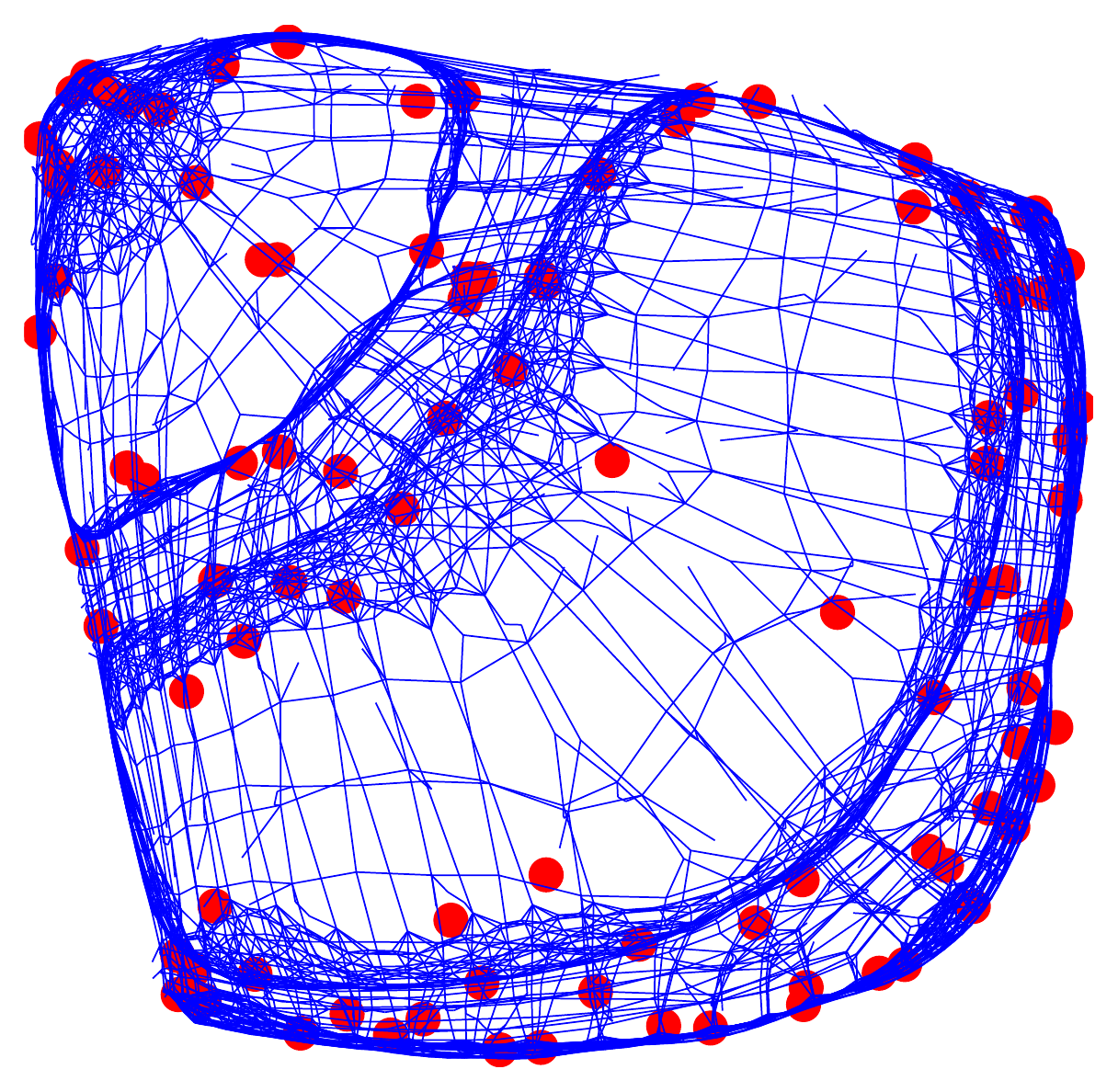}}
		\centerline{vp-100}
	\end{minipage}
	\begin{minipage}{0.16\linewidth}
		\vspace{3pt}
		\centerline{\includegraphics[width=1.3in, angle=-180]{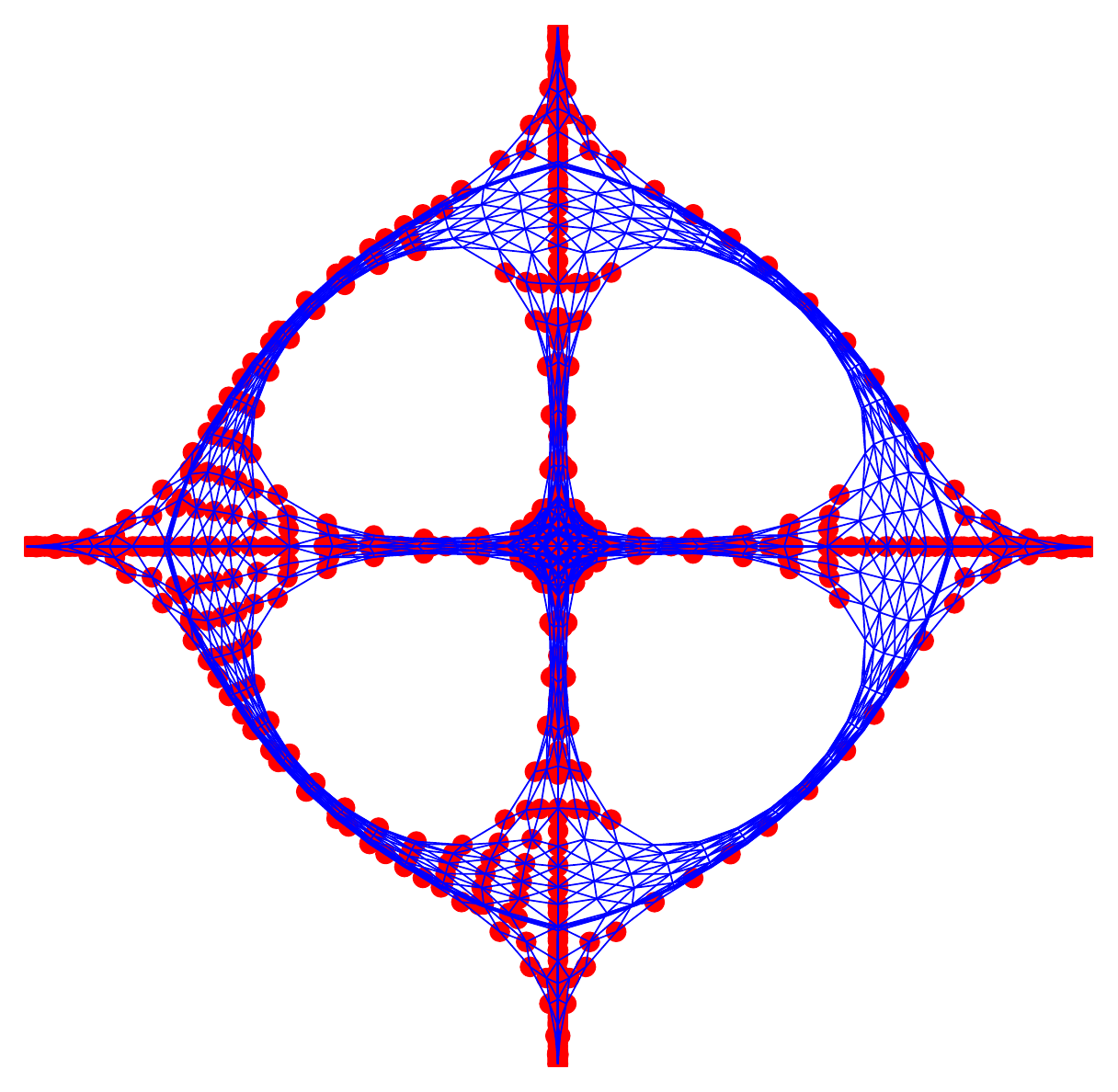}}
		\vspace{3pt}
 		\centerline{\includegraphics[width=1.1in]{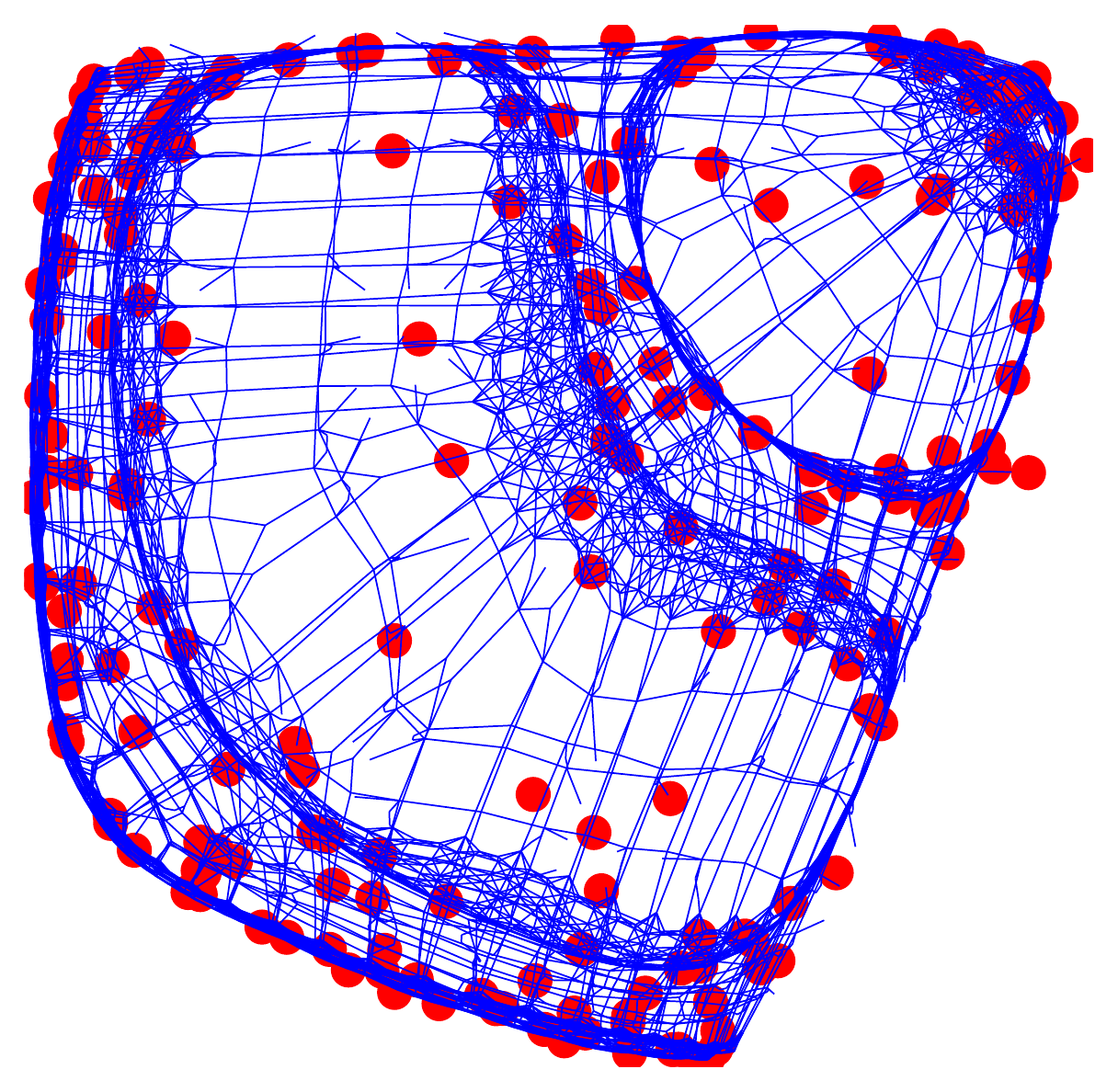}}
		\centerline{vp-200}
	\end{minipage}

	\caption{B-Stress graph drawing of pivot-enhanced graphs. For clarity, virtual edges are not drawn, and red nodes represent pivots.}
	\label{Fig.10}
\end{figure*}

% \usepackage{multirow}

\renewcommand{\thetable}{\arabic{table}}
\renewcommand{\thetable}{11}

\begin{table*}[!t]
\caption{The performance of GCN on pivot-augmented graphs.}
\label{table-11}
\centering

\setlength{\tabcolsep}{1.6mm}{
\begin{tabular}{ccccccccc}
\hline
\textbf{Dataset}               & \textbf{GCN-VP0}                                        & \textbf{GCN-VP1}                             & \textbf{GCN-VP2}                    & \textbf{GCN-VP3}                             & \textbf{GCN-VP5}                             & \textbf{GCN-VP7}                             & \textbf{GCN-VP10}                            & \textbf{GCN-VP20}                            \\ \hline
                               & \cellcolor[HTML]{FFFA70}0.767±0.027                     & \cellcolor[HTML]{FFFA70}0.734±0.038          & \cellcolor[HTML]{FFFA70}0.740±0.035 & \cellcolor[HTML]{FFFA70}0.762±0.031          & \cellcolor[HTML]{FFFA70}0.759±0.026          & \cellcolor[HTML]{FFFA70}0.732±0.008          & \cellcolor[HTML]{FFFA70}\textbf{0.796±0.015} & \cellcolor[HTML]{FFFA70}\textbf{0.795±0.013} \\
                               & \cellcolor[HTML]{FFFA70}GIN-VP0                         & \cellcolor[HTML]{FFFA70}GIN-VP1              & \cellcolor[HTML]{FFFA70}GIN-VP2     & \cellcolor[HTML]{FFFA70}GIN-VP3              & \cellcolor[HTML]{FFFA70}GIN-VP5              & \cellcolor[HTML]{FFFA70}GIN-VP7              & \cellcolor[HTML]{FFFA70}GIN-VP10             & \cellcolor[HTML]{FFFA70}GIN-VP20             \\
\multirow{-3}{*}{Ogbg-molbace} & \cellcolor[HTML]{FFFA70}0.725±0.025                     & \cellcolor[HTML]{FFFA70}0.780±0.025          & \cellcolor[HTML]{FFFA70}0.749±0.007 & \cellcolor[HTML]{FFFA70}0.767±0.038          & \cellcolor[HTML]{FFFA70}0.766±0.019          & \cellcolor[HTML]{FFFA70}0.787±0.033          & \cellcolor[HTML]{FFFA70}\textbf{0.809±0.020} & \cellcolor[HTML]{FFFA70}\textbf{0.813±0.020} \\
                               & \cellcolor[HTML]{22D2CA}GCN-VP0                         & \cellcolor[HTML]{22D2CA}GCN-VP1              & \cellcolor[HTML]{22D2CA}GCN-VP2     & \cellcolor[HTML]{22D2CA}GCN-VP3              & \cellcolor[HTML]{22D2CA}GCN-VP5              & \cellcolor[HTML]{22D2CA}GCN-VP7              & \cellcolor[HTML]{22D2CA}GCN-VP10             & \cellcolor[HTML]{22D2CA}GCN-VP20             \\
                               & \cellcolor[HTML]{22D2CA}0.642±0.012                     & \cellcolor[HTML]{22D2CA}\textbf{0.687±0.012} & \cellcolor[HTML]{22D2CA}0.679±0.011 & \cellcolor[HTML]{22D2CA}0.677±0.005          & \cellcolor[HTML]{22D2CA}0.662±0.016          & \cellcolor[HTML]{22D2CA}\textbf{0.683±0.010} & \cellcolor[HTML]{22D2CA}\textbf{0.691±0.011} & \cellcolor[HTML]{22D2CA}\textbf{0.684±0.007} \\
                               & \cellcolor[HTML]{22D2CA}GIN-VP0                         & \cellcolor[HTML]{22D2CA}GIN-VP1              & \cellcolor[HTML]{22D2CA}GIN-VP2     & \cellcolor[HTML]{22D2CA}GIN-VP3              & \cellcolor[HTML]{22D2CA}GIN-VP5              & \cellcolor[HTML]{22D2CA}GIN-VP7              & \cellcolor[HTML]{22D2CA}GIN-VP10             & \cellcolor[HTML]{22D2CA}GIN-VP20             \\
\multirow{-4}{*}{Ogbg-molbbbp} & \multicolumn{1}{l}{\cellcolor[HTML]{22D2CA}0.671±0.025} & \cellcolor[HTML]{22D2CA}0.671±0.015          & \cellcolor[HTML]{22D2CA}0.676±0.012 & \cellcolor[HTML]{22D2CA}\textbf{0.681±0.011} & \cellcolor[HTML]{22D2CA}\textbf{0.682±0.013} & \cellcolor[HTML]{22D2CA}\textbf{0.710±0.015} & \cellcolor[HTML]{22D2CA}\textbf{0.697±0.016} & \cellcolor[HTML]{22D2CA}\textbf{68.7±0.09}   \\ \hline
\end{tabular}
}
\end{table*}

% \usepackage{multirow}
\renewcommand{\thetable}{\arabic{table}}
\renewcommand{\thetable}{10}
\begin{table}
\caption{Pivot tests against over-smoothing.}
\label{table-10}
\centering

\setlength{\tabcolsep}{1.99mm}{
\begin{tabular}{ll|ccccc}
\hline
                                                &                                   & \multicolumn{5}{c}{\textbf{Layers}}                                                                                                                                                      \\
\multirow{-2}{*}{\textbf{Dataset}}              & \multirow{-2}{*}{\textbf{Model}}  & \textbf{2}                             & \textbf{4}                    & \textbf{8}                             & \textbf{16}                   & \textbf{32}                            \\ \hline
                                                & \cellcolor[HTML]{FFFA70}GCN       & \cellcolor[HTML]{FFFA70}86.27          & \cellcolor[HTML]{FFFA70}85.49 & \cellcolor[HTML]{FFFA70}84.31          & \cellcolor[HTML]{FFFA70}7.84  & \cellcolor[HTML]{FFFA70}7.84           \\
                                                & \cellcolor[HTML]{FFFA70}GCN-VP(1) & \cellcolor[HTML]{FFFA70}87.84          & \cellcolor[HTML]{FFFA70}83.14 & \cellcolor[HTML]{FFFA70}80.78          & \cellcolor[HTML]{FFFA70}68.24 & \cellcolor[HTML]{FFFA70}53.3           \\
                                                & \cellcolor[HTML]{FFFA70}GCN-VP(5) & \cellcolor[HTML]{FFFA70}\textbf{88.24} & \cellcolor[HTML]{FFFA70}87.06 & \cellcolor[HTML]{FFFA70}82.35          & \cellcolor[HTML]{FFFA70}70.59 & \cellcolor[HTML]{FFFA70}61.18          \\
\multirow{-4}{*}{MSRC\_9}                       & \cellcolor[HTML]{FFFA70}StressGCN & \cellcolor[HTML]{FFFA70}\textbf{88.24} & \cellcolor[HTML]{FFFA70}86.27 & \cellcolor[HTML]{FFFA70}86.67          & \cellcolor[HTML]{FFFA70}87.45 & \cellcolor[HTML]{FFFA70}88.24          \\
\multicolumn{1}{c}{}                            & \cellcolor[HTML]{22D2CA}GCN       & \cellcolor[HTML]{22D2CA}67.18          & \cellcolor[HTML]{22D2CA}63.59 & \cellcolor[HTML]{22D2CA}59.49          & \cellcolor[HTML]{22D2CA}23.08 & \cellcolor[HTML]{22D2CA}10.26          \\
\multicolumn{1}{c}{}                            & \cellcolor[HTML]{22D2CA}GCN-VP(1) & \cellcolor[HTML]{22D2CA}\textbf{69.23} & \cellcolor[HTML]{22D2CA}66.67 & \cellcolor[HTML]{22D2CA}58.46          & \cellcolor[HTML]{22D2CA}33.85 & \cellcolor[HTML]{22D2CA}31.62          \\
\multicolumn{1}{c}{}                            & \cellcolor[HTML]{22D2CA}GCN-VP(5) & \cellcolor[HTML]{22D2CA}62.56          & \cellcolor[HTML]{22D2CA}68.21 & \cellcolor[HTML]{22D2CA}62.05          & \cellcolor[HTML]{22D2CA}35.38 & \cellcolor[HTML]{22D2CA}31.28          \\
\multicolumn{1}{c}{\multirow{-4}{*}{MSRC\_21C}} & \cellcolor[HTML]{22D2CA}StressGCN & \cellcolor[HTML]{22D2CA}\textbf{71.28} & \cellcolor[HTML]{22D2CA}65.13 & \cellcolor[HTML]{22D2CA}67.69          & \cellcolor[HTML]{22D2CA}70.26 & \cellcolor[HTML]{22D2CA}71.28          \\
                                                & \cellcolor[HTML]{FFCCC9}GCN       & \cellcolor[HTML]{FFCCC9}63.71          & \cellcolor[HTML]{FFCCC9}64.74 & \cellcolor[HTML]{FFCCC9}56.08          & \cellcolor[HTML]{FFCCC9}1.24  & \cellcolor[HTML]{FFCCC9}1.03           \\
                                                & \cellcolor[HTML]{FFCCC9}GCN-VP(1) & \cellcolor[HTML]{FFCCC9}62.27          & \cellcolor[HTML]{FFCCC9}74.02 & \cellcolor[HTML]{FFCCC9}\textbf{74.43} & \cellcolor[HTML]{FFCCC9}21.44 & \cellcolor[HTML]{FFCCC9}12.99          \\
                                                & \cellcolor[HTML]{FFCCC9}GCN-VP(5) & \cellcolor[HTML]{FFCCC9}65.77          & \cellcolor[HTML]{FFCCC9}71.13 & \cellcolor[HTML]{FFCCC9}59.79          & \cellcolor[HTML]{FFCCC9}23.51 & \cellcolor[HTML]{FFCCC9}17.11          \\
\multirow{-4}{*}{Cuneiform}                     & \cellcolor[HTML]{FFCCC9}StressGCN & \cellcolor[HTML]{FFCCC9}62.68          & \cellcolor[HTML]{FFCCC9}63.30 & \cellcolor[HTML]{FFCCC9}62.47          & \cellcolor[HTML]{FFCCC9}63.51 & \cellcolor[HTML]{FFCCC9}\textbf{64.33} \\ \hline
\end{tabular}
}
\end{table}

\vspace{0.3cm}
\noindent \textit{4.4.2 Analysis and Verification of Stress Virtual Pivot GNNs}
\vspace{0.1cm}

Although we can continuously reduce the complexity of the distance calculation in above repulsive models, combining attractive and repulsive information is still a challenge. Adding virtual pivots to the graph, without designating a different target distance for each node pair, can serve as a compromise solution to approximate the full stress message propagation. For virtual pivot GNNs, we report on the following two sets of experiments.

\textbf{Over-smoothing test}. Table 10 summarizes the graph classification results for virtual pivot models with various numbers of layers. Compared with GCN, adding one (GCN-VP(1)) or five pivots (GCN-VP(5)) is a simple but powerful trick. They can alleviate over-smoothing to some extent. On the MSRC\_9 and CUNeiform datasets, they even achieve similar or even better results than StressGCN. In theory, the virtual pivot-based GNNs can not prevent over-smoothing since most of the non-neighboring pairwise distances are missing. Maintaining the node spacing can prevent nodes from collapsing into the same location but may not form or maintain an effective group structure. We can only seek a subtle balance between the simplified repulsion and local message passing.

\textbf{Multi-pivot test}. We experiment on two OGB datasets to assess how the number of virtual pivots will affect the model performance. As shown in Table 11, the number of pivots gradually increases from 1 to 20. We can see a clear trend that the impact of virtual nodes grows when the number of pivots increases. Slightly different from the case of repulsive message passing, multiple pivots split the whole graph into different regions and provide a shortcut for sharing information inside each region. Therefore, it is reasonable to conclude that virtual pivots could merge local and repulsive message-passing advantages.

Moreover, we are also concerned with the evaluation of the impact of virtual pivots in graph visualization. In Section 3, from the perspective of global distance, we expound the equivalent relationship between the underlying graph of B-Stress and the graph with a single virtual node. It is easy to infer that one virtual node has no impact on the layout of the B-Stress graph drawing. What about multiple pivots? We here answer in the view of the B-Stress graph visualization. In Fig. 10, we varied the number of pivots from 0, 1, 5, 10, 100 to 200. For one virtual pivot, the original graph structure is clear and may be slightly deformed (vp-1 v.s. vp-0). When increasing the number of pivots to 5 and 10, these virtual points stretch and distort the entire graph. However, the density of nodes or edges is still relatively maintained. For example, for the two images in the middle of the first row of Fig. 10, we can still recognize 5 groups. When setting the number of pivots to 100 or 200, the resulting graph layout is equivalent to the B-Stress layout of the original graph (vp-100, vp-200 v.s. vp-0). Therefore, when the number of virtual pivots is large enough, the added virtual points have a negligible effect on graph visualization.

\section{Conclusion}

 In the process of designing and applying graph neural networks, we often fall into some optimization pitfalls, such as overemphasizing the antagonistic relationship between over-smoothing and deep models, blindly deepening models, blindly migrating popular techniques and concepts from deep learning to GNNs, and wrongly attributing the effectiveness of the model to unnecessary components. The main cause is that we do not understand the mechanism of graph neural networks. Although knowledge from deep learning has enabled the graph learning community to make substantial gains in recent years, it is likely to impede progress in the long term. To the best of our knowledge, stress graph drawing is the best resource for understanding and designing graph neural networks. It allows us to overcome optimization pitfalls from which previous methods suffer. It also gives us new directions for optimizing current GNNs, that is, how to use repulsive information and how to better simplify the full stress message iteration. Existing theories of GNNs can not provide such a unique viewpoint. We believe that our stress graph neural networks will become a popular GNN framework.
 
\section*{Acknowledgement}

We thank Yifan Hu (AT\&T Labs Research) for providing the Barnes \& Hut algorithm on his GitHub and the graph visualization data in the SuiteSparse Matrix Collection. The author, Xue Li, would like to express his sincere gratitude to Yehuda Koren (AT\&T Lab Research), whose work inspires and encourages him a lot. Finally, Xue Li wants to thank, in particular, the invaluable love and support from Sidan Wang over the years. Will you marry me?

\ifCLASSOPTIONcaptionsoff
  \newpage
\fi



%\bibliographystyle{IEEEtran}


\vspace{0.3cm}
\noindent \textbf{CRediT authorship contribution statement}
\vspace{0.1cm}

\textbf{Xue Li}: Conceptualization, Data curation, Methodology, Writing-original draft. \textbf{Yuanzhi Cheng}: Supervision, Writing–review \& editing.

\begin{IEEEbiography}[{\includegraphics[width=1in,height=1.25in,clip,keepaspectratio]{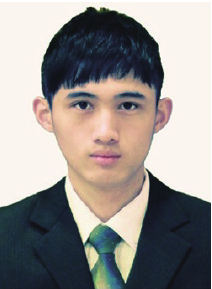}}]{Xue Li}
is a Ph.D. student in the School of Computer Science and Technology, Harbin Institute of Technology. His research interests are concentrated on the interpretation of graph neural networks and popular deep learning techniques, graph visualization, and randomized algorithms.
\end{IEEEbiography}

\begin{IEEEbiography}[{\includegraphics[width=1in,height=1.25in,clip,keepaspectratio]{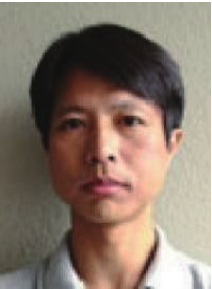}}]{Yuanzhi Cheng}
received his Ph.D. through the joint training of Osaka University and Harbin Institute of Technology in 2007. He is currently a Professor at the Harbin Institute of technology until 2019. His research interests are concentrated on machine learning, image processing, and computer vision. He has published over 50 journal papers in IEEE Trans. Image. Process, IEEE Trans. Biomed. Eng, IEEE J. Biomed. Health Inform, Med. Image Anal, Pattern Recognition, etc.
\end{IEEEbiography}


\vfill


\end{document}